\documentclass[journal,twoside,web]{ieeecolor}

\usepackage{cite}

\usepackage{amsmath,amssymb,amsfonts}
\usepackage{algorithmic}
\usepackage{graphicx}
\usepackage{algorithm,algorithmic}
\usepackage{generic}
\usepackage[hidelinks]{hyperref}
\usepackage{textcomp}

\def\BibTeX{{\rm B\kern-.05em{\sc i\kern-.025em b}\kern-.08em
    T\kern-.1667em\lower.7ex\hbox{E}\kern-.125emX}}

\begin{document}
\title{On the State of NLP Approaches to Modeling Depression in Social Media: A Post-COVID-19 Outlook}

\author{Ana-Maria Bucur, Andreea-Codrina Moldovan, Krutika Parvatikar, Marcos Zampieri,\\ Ashiqur R. KhudaBukhsh and Liviu P. Dinu
\thanks{This research is supported by the project “Romanian Hub for Artificial Intelligence - HRIA”, Smart Growth, Digitization, and Financial Instruments Program, 2021-2027, MySMIS no. 334906.}
\thanks{Ana-Maria Bucur is with the Interdisciplinary School of Doctoral Studies, University of Bucharest, Bucharest 050663, Romania and PRHLT Research Center, Universitat Politècnica de València, València 46020, Spain (e-mail: ana-maria.bucur@drd.unibuc.ro). }
\thanks{Andreea-Codrina Moldovan is with the Interdisciplinary School of Doctoral Studies, University of Bucharest, Bucharest 050663, Romania (e-mail: moldovanandreeacodrina@gmail.com).}
\thanks{Krutika Parvatikar and Ashiqur R. KhudaBukhsh are with Rochester Institute of Technology, New York 14623, USA (e-mails: kp4365@rit.edu, axkvse@rit.edu).}
\thanks{Marcos Zampieri is with George Mason University, Virginia 22030, USA (e-mail: mzampier@gmu.edu).}
\thanks{Liviu P. Dinu is with the Human Language Technology Research Center and Department of Computer Science, Faculty of Mathematics and Computer Science, University of Bucharest, Bucharest 050663, Romania
(e-mail: ldinu@fmi.unibuc.ro).}}

\maketitle

\begin{abstract}
Computational approaches to predicting mental health conditions in social media have been substantially explored in the past years. Multiple reviews have been published on this topic, providing the community with comprehensive accounts of the research in this area. Among all mental health conditions, depression is the most widely studied due to its worldwide prevalence. The COVID-19 global pandemic, starting in early 2020, has had a great impact on mental health worldwide. Harsh measures employed by governments to slow the spread of the virus (e.g., lockdowns) and the subsequent economic downturn experienced in many countries have significantly impacted people's lives and mental health. Studies have shown a substantial increase of above 50\% in the rate of depression in the population. In this context, we present a review on natural language processing (NLP) approaches to modeling depression in social media, providing the reader with a post-COVID-19 outlook. This review contributes to the understanding of the impacts of the pandemic on modeling depression in social media. We outline how state-of-the-art approaches and new datasets have been used in the context of the COVID-19 pandemic. Finally, we also discuss ethical issues in collecting and processing mental health data, considering fairness, accountability, and ethics.
\end{abstract}

\begin{IEEEkeywords}
COVID-19, Depression, NLP, Mental health, Social media
% Enter key words or phrases in alphabetical order, separated by commas. Using the IEEE Thesaurus can help you find the best standardized keywords to fit your article. Use the thesaurus access request form for free access to the IEEE Thesaurus: \underline{https://www.ieee.org/publications/services/thesaurus-acce}\\
% \underline{ss-page.com.}
\end{IEEEkeywords}

\section{Introduction}
The early identification and diagnosis of depression and other mental health conditions are vital for effective early treatment of such diseases and for mitigating their negative effects. Early interventions have proven to significantly reduce the impact of mental disorders such as depression, bipolar disorders, eating disorders, and many others \cite{davey2019early,read2022history}. In this context, the vast amount of user-generated content available on social media allows researchers to use computational models to investigate signs and symptoms of various mental health conditions using online data. Studies have shown encouraging results in predicting various mental disorders using social media analysis and natural language processing (NLP) models \cite{chancellor2020methods}. Social media platforms have become a valuable source of data, enabling low-cost and non-invasive early interventions.

Depression is one of the mental health conditions most widely studied using social media \cite{chancellor2020methods}. According to data from the World Health Organization (WHO)\footnote{\href{https://www.who.int/news-room/fact-sheets/detail/depression}{https://www.who.int/news-room/fact-sheets/detail/depression}}, depression is also the most prevalent mental health condition. It is estimated that nearly 4\% of the world's population suffers from depression, including 5\% of adults and almost 6\% of individuals aged 60 or older. Mild and moderate cases of depression lead to negative consequences to someone's well-being, while more severe cases of depression can ultimately lead to suicide. 

Starting on March 11\textsuperscript{th}, 2020, the COVID-19 global pandemic brought a number of negative social and economic consequences with serious repercussions on mental health all over the world. Prolonged social isolation and fear of contagion combined with harsh and unprecedented government measures such as school and business closures have created the conditions for an anxiety and depression epidemic. WHO estimates that the prevalence of anxiety and depression worldwide has spiked by around 25\% in the first year of the pandemic alone.\footnote{\href{https://www.who.int/news/item/02-03-2022-covid-19-pandemic-triggers-25-increase-in-prevalence-of-anxiety-and-depression-worldwide}{https://www.who.int/news/item/02-03-2022-covid-19-pandemic-triggers-25-increase-in-prevalence-of-anxiety-and-depression-worldwide}} Studies on the prevalence of depression published over the course of the pandemic estimate even higher increase rates of depression of above 50\% \cite{wolohan2020estimating,kaseb2022analysis}. On May 5\textsuperscript{th}, 2023, after more than three years since the beginning of the pandemic, the WHO announced that COVID-19 is no longer ``a public health emergency of international concern'' \cite{lenharo2023declares}. The WHO announcement marked, at least in practice, the end of the COVID-19 pandemic, but the negative effects of the global public health emergency are likely to continue in the years to come. 

In this paper, we present the first review of NLP approaches developed to model depression in social media with a post-COVID-19 pandemic outlook. We present datasets and models used within the context of this public health emergency. To the best of our knowledge, this is the first review conducted after the end of the pandemic providing the reader with a post-COVID-19 outlook on how the pandemic impacted the mental health of the global population and how this influenced research on depression using social media analysis and NLP. In this review, we focus on depression defined as major depressive disorder (MDD)\footnote{\href{https://www.psychiatry.org/patients-families/depression/what-is-depression}{https://www.psychiatry.org/patients-families/depression/what-is-depression}} with the code F33 in the ICD-10 taxonomy.\footnote{\href{https://www.icd10data.com/ICD10CM/Codes/F01-F99/F30-F39/F33-}{https://www.icd10data.com/ICD10CM/Codes/F01-F99/F30-F39/F33-}} Other often related mood disorders, such as bipolar disorder, peripartum depression, seasonal affective disorder are outside the scope of this review.

\subsection{Past Related Reviews}

As discussed in this work, the interest in modeling mental health on social media continues to grow and this is reflected in the number of papers published in this area. Since the review by Calvo et al. (2017) \cite{calvo2017natural}, there have been multiple reviews covering topics at the intersection of mental health, social media, and NLP. We present related reviews below, ordered by the year of publication.\footnote{Most papers listed in this section are referred to as \emph{surveys} within computer science and related fields. To avoid confusion with survey papers in the medical domain, which entail questionnaire studies, we refer to the papers listed here as \emph{reviews}.}

\begin{itemize}
    \item Calvo et al. (2017) \cite{calvo2017natural} is arguably the first comprehensive review at the intersection of mental health and NLP. The paper considers work on various mental health conditions and states such as depression, mood disorders, psychological distress, and suicide ideation in \emph{non-clinical texts}. Under \emph{non-clinical texts}, the authors include work on data retrieved from social media platforms and other forms of user-generated content online, such as live journals and online forums. The review also includes work on suicide notes.
    \item The review by Skaik and Inkpen (2020) \cite{skaik2020using} presents studies applying machine learning (ML) and NLP to model various mental health conditions. It focuses on papers published between 2013 and 2020 on English social media data.
    \item Chancellor and De Choudhury (2020) \cite{chancellor2020methods} focus on predicting mental health status on social media published between 2013 and 2018. The authors present a critical review of the papers included and provide some recommendations to improve the methods used for research in the field.
    \item R{\'\i}ssola et al. (2021) \cite{rissola2021survey} considers papers on online mental state assessment using social media, a similar focus to the review by Chancellor and De Choudhury (2020) \cite{chancellor2020methods}. A novel aspect addressed by  R{\'\i}ssola et al. (2021) \cite{rissola2021survey} and not captured by previous reviews is the discussion of insights on language and behavior obtained by the use of computational models on social media data. 
    \item Zhang et al. (2022) \cite{zhang2022natural} focus on mental health illness detection instead of mental states, and is one of the most comprehensive, covering a total of nearly 400 studies on depression, bipolar disorder, schizophrenia, etc.
    \item Dhelim et al. (2023) \cite{dhelim2023detecting} is the most recent related review. It is also the only review to focus on COVID-19. It focuses on general mental well-being, including loneliness, anxiety, stress, post-traumatic stress disorder (PTSD), and other mental disorders, in addition to depression and suicide ideation.
\end{itemize}

\noindent Our review is unique in its scope as it provides the community with a post-pandemic outlook on modeling depression in social media using NLP. We focus on a single mental health condition, depression, which is the most prevalent mental health condition worldwide and the one that has been mostly studied within the context of the COVID-19 pandemic. Compared to the work of Dhelim et al. (2023) \cite{dhelim2023detecting}, we provide a comprehensive overview of the recent articles published during the pandemic, focusing on how depression evolved during this time. We believe that the post-pandemic outlook is important and presents interesting opportunities for future research.

\subsection{Roadmap}

The remainder of this paper is organized as follows. Section \ref{sec:depression} presents an overview of NLP approaches applied to depression detection on social media, from early methods that rely on feature engineering and classic machine learning classifiers to state-of-the-art approaches based on word embeddings and neural networks. This section also discusses international workshops and benchmark competitions organized on this topic, such as CLPsych and eRisk. Section \ref{sec:methodology} outlines the methodology used for conducting this review. Section \ref{sec:covid-19} focuses on the COVID-19 pandemic. We look at how the COVID-19 pandemic impacted the mental health of the general population and healthcare workers, and how research on modeling using NLP and social media analysis was applied in this context. Section \ref{sec:datasets} presents the pandemic-related data collections. Section \ref{sec:discussion} discusses ethical considerations when working on this topic as well as questions related to fairness and bias. Finally, Section \ref{sec:conclusion} concludes this review and presents avenues for future research.

\section{Depression modeling}
\label{sec:depression}
The field of modeling mental disorders from social media data using NLP approaches has seen a growing interest over time, driven by the organization of workshops and shared tasks specifically focused on mental disorders, such as:

\vspace{1mm}
\noindent \textbf{eRisk~~} The Early Risk Prediction on the Internet Lab\footnote{\href{https://erisk.irlab.org/}{https://erisk.irlab.org/}} \cite{losada2016test} started in 2017 with a pilot task on depression detection from textual social media data. Since then, the lab has consistently organized several shared tasks each year. Moreover, the eRisk Lab has expanded its scope to other mental health problems: self-harm, eating disorders, or pathological gambling. 

\vspace{1mm}
\noindent \textbf{CLPsych~~} The Workshop on Computational Linguistics and Clinical Psychology\footnote{\href{https://clpsych.org/}{https://clpsych.org/}} started in 2014, and over time has organized various shared tasks on depression and PTSD \cite{coppersmith2015clpsych}, suicide risk \cite{zirikly2019clpsych}, and predicting current and future psychological health from childhood essays \cite{lynn2018clpsych}.

\vspace{1mm}
\noindent \textbf{LT-EDI~~} The Workshop on Language Technology for Equality, Diversity, and Inclusion\footnote{\href{https://sites.google.com/view/lt-edi-2023}{https://sites.google.com/view/lt-edi-2023}} \cite{kayalvizhi2022findings,s-etal-2023-overview-shared} has also organized two shared tasks on detecting depression signs from social media data in 2022 and 2023.

\vspace{1mm}
\noindent \textbf{MentalRiskES~~} The IberLEF 2023 shared evaluation campaign hosted the Early Detection of Mental Disorders Risk in Spanish\footnote{\href{https://sites.google.com/view/mentalriskes}{https://sites.google.com/view/mentalriskes}} \cite{marmol2023overview} task. It was the first venue to propose a shared task on detecting mental disorders in languages other than English. MentalRiskES hosted three different tasks on detecting eating disorders, depression, and anxiety.

These workshops and shared tasks have been valuable in providing data for research and fostering the development of novel approaches for depression modeling. This research has paved the way for the social media analyses and methods that were used to asses the impact of the COVID-19 pandemic.

\subsection{Early Approaches}

Most previous works model the task as a binary classification, where the goal is to predict a label for each data point, such as depressed or non-depressed. For such tasks, traditional machine learning algorithms were consistently used: Logistic Regression \cite{cohan2018smhd,Eichstaedt11203}, SVM \cite{jamil-etal-2017-monitoring,Yazdavar2020MultimodalMH}, Naive Bayes \cite{ijcai2017p536, ALSAGRI_2020,article23}, Random Forest \cite{bucur2021psychologically, birnbaum2020identifying}, Decision Trees \cite{info:doi/10.2196/17650, inproceedingsvedula}. These classical machine learning classifiers often made use of several language markers uncovered in psychology research, such as the use of first-person pronouns ``I" and ``we" \cite{rude2004language,bucur2021psychologically,jamil-etal-2017-monitoring}, of negative or offensive terms \cite{fekete2002internet,bucur2021exploratory}, or emotional patterns \cite{inproceedingsvedula,article23}. Classic machine learning classifiers have been used extensively not only for depression detection, but also for various other mental disorders, such as anxiety and schizophrenia, which are not the focus of our review.

\subsection{State-of-the-art}

Due to computational advances in the machine learning field and the rise in popularity of deep neural network architectures, approaches such as CNNs \cite{yates-etal-2017-depression,hamad2021depressionnet}, RNNs \cite{10.1007/s10844-020-00599-5,suhara2017deepmood}, LSTMs \cite{nanomi-arachchige-etal-2021-dataset-research,sadeque2018measuring} and MLPs \cite{nanomi-arachchige-etal-2021-dataset-research} have been used for detecting the presence of mental disorders signs in online data. Deep learning models based on pre-trained transformer architectures such as BERT \cite{haque2021deep,bucur2021early,wolk2021hybrid,podina14mental}, XLNet \cite{dinu2021automatic,DBLP:journals/corr/abs-2102-09427} and RoBERTa \cite{Wu2021ARM,bucur2022end} are often used for modeling depression. 
%Unlike traditional text representation methods such as Bag-of-Words (BOW) and Term frequency inverse-document frequency (TF-IDF), BERT \cite{kenton2019bert} provides sentence or document representations that contain context information. Hence, the model can yield a more genuine encapsulation of meaning. RoBERTa, a more stable approach to BERT, solves the latter's undertraining problem. XLNet \cite{DBLP:journals/corr/abs-1906-08237}, a similar model using the bidirectional encoder, manages to overrun BERT using an autoregressive training method. 

Pre-trained models specific to particular domains are becoming more popular, alongside transformer models, due to their effectiveness and state-of-the-art results on downstream tasks. MentalBERT and MentalRoBERTa \cite{ji2022mentalbert}, which are specialized variants of BERT and RoBERTa models, have shown promising results in identifying language patterns associated with mental health problems. In addition, MentalLongformer and MentalXLNet \cite{ji2023domain} were proposed to capture the long context of discourse related to mental health issues. Naseem et al. (2022) \cite{naseem2022benchmarking} proposed PHS-BERT, a model trained on health-related tweets that can be used for public health surveillance on social media data. Another pre-trained model, DisorBERT \cite{aragon2023disorbert}, was proposed to detect signs of mental disorders (depression, self-harm and depression) in social media data.

Recent advances in large language models (LLMs) have opened up potential applications for depression modeling. These applications include the assessment of symptoms of mental disorders \cite{xu2023leveraging,yang2023evaluations,qin2023read,amin38will,perez2023depresym} and the generation of synthetic data \cite{bucur2023utilizing}. 
Moreover, LLMs have shown great potential for explainable detection of depression, generating explanations that can be used by professionals to interpret the predictions of the models \cite{yang2023evaluations,souto2023explainability,yang2023mentalllama}. Xu et al. (2023) \cite{xu2023leveraging} evaluated multiple LLMs, such as Alpaca \cite{taori2023stanford}, FLAN-T5 \cite{chung2022scaling}, GPT-3.5 \cite{openai2022chatgpt} and GPT-4 \cite{bubeck2023sparks}, on multiple tasks on mental disorders identification. In zero-shot and few-shot prompt settings, the models showed limited performance. However, using instruction fine-tuning, the performance on all tasks improved simultaneously. Souto et al. (2023) \cite{souto2023explainability} has also shown that fine-tuned pre-trained language models (e.g., BERT, MentalBERT) outperform LLMs such as GPT 3.5 \cite{openai2022chatgpt} and Vicuna \cite{chiang2023vicuna} for depression classification. To address the limitations of general-purpose LLMs, Yang et al. (2023) \cite{yang2023mentalllama} proposed MentaLLaMA, an open-source instruction-tuned model for mental health analysis, which showed greater performance than general-purpose LLMs.

\section{Methodology}
\label{sec:methodology}
In this section, we outline the methodology for conducting our review of the NLP approaches to modeling depression through social media data during the pandemic. The aim of this review is to provide an overview of the methods employed to monitor social media and offer the reader a post-COVID-19 outlook.

We have conducted a comprehensive literature search on the major publication databases, including ACL Anthology, IEEE Xplore, Scopus, ACM Digital Library, Springer Nature Link, ScienceDirect, and Google Scholar. We formulated the following search query to retrieve relevant papers:

\noindent (“depression” OR “depression detection” OR “depression prediction” OR “depression monitoring”) AND “social media” AND (“covid” OR "pandemic" OR “covid-19 pandemic”)

For this review, we have selected papers published during the pandemic, between 2020 and 2023, specifically focusing on studies that conduct depression-related analyses in the context of the COVID-19 pandemic using social media data. Papers that were not written in English were excluded. To ensure relevance, one of the key criteria for inclusion was the focus on COVID-19. All full texts of the retrieved papers were manually inspected to determine whether they were performing depression-related analyses on social media data posted during the COVID-19 pandemic. From a total of 231 papers identified through our database search, only 38 were focused on depression modeling using social media data posted during the pandemic.

\section{Depression modeling during COVID-19 pandemic}
\label{sec:covid-19}
\begin{figure}[!hbt]
	\centering
		\includegraphics[width=.35\textwidth]{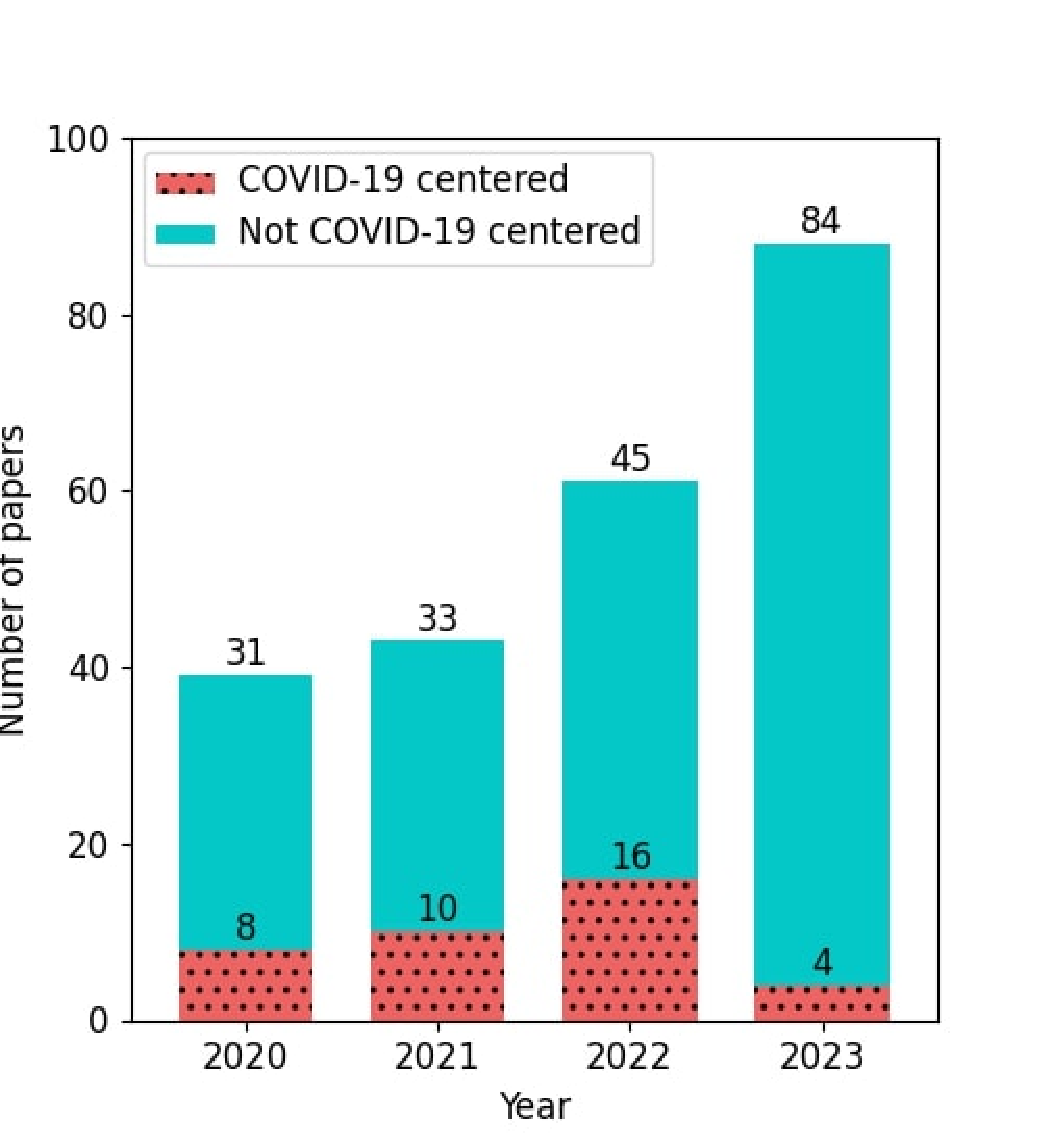}
	\caption{Number of papers on depression modeling published during the pandemic (2020-2023) in peer-reviewed conferences or journals.}
	\label{num-of-papers-depression}
\end{figure}

This section provides a detailed overview of the studies gathered for this review. In Figure \ref{num-of-papers-depression}, we show the number of papers on depression modeling from social media data published every year since the beginning of the COVID-19  pandemic in 2020 until the end of the pandemic in 2023.

\begin{figure}[!hbt]
	\centering
		\includegraphics[width=.48\textwidth]{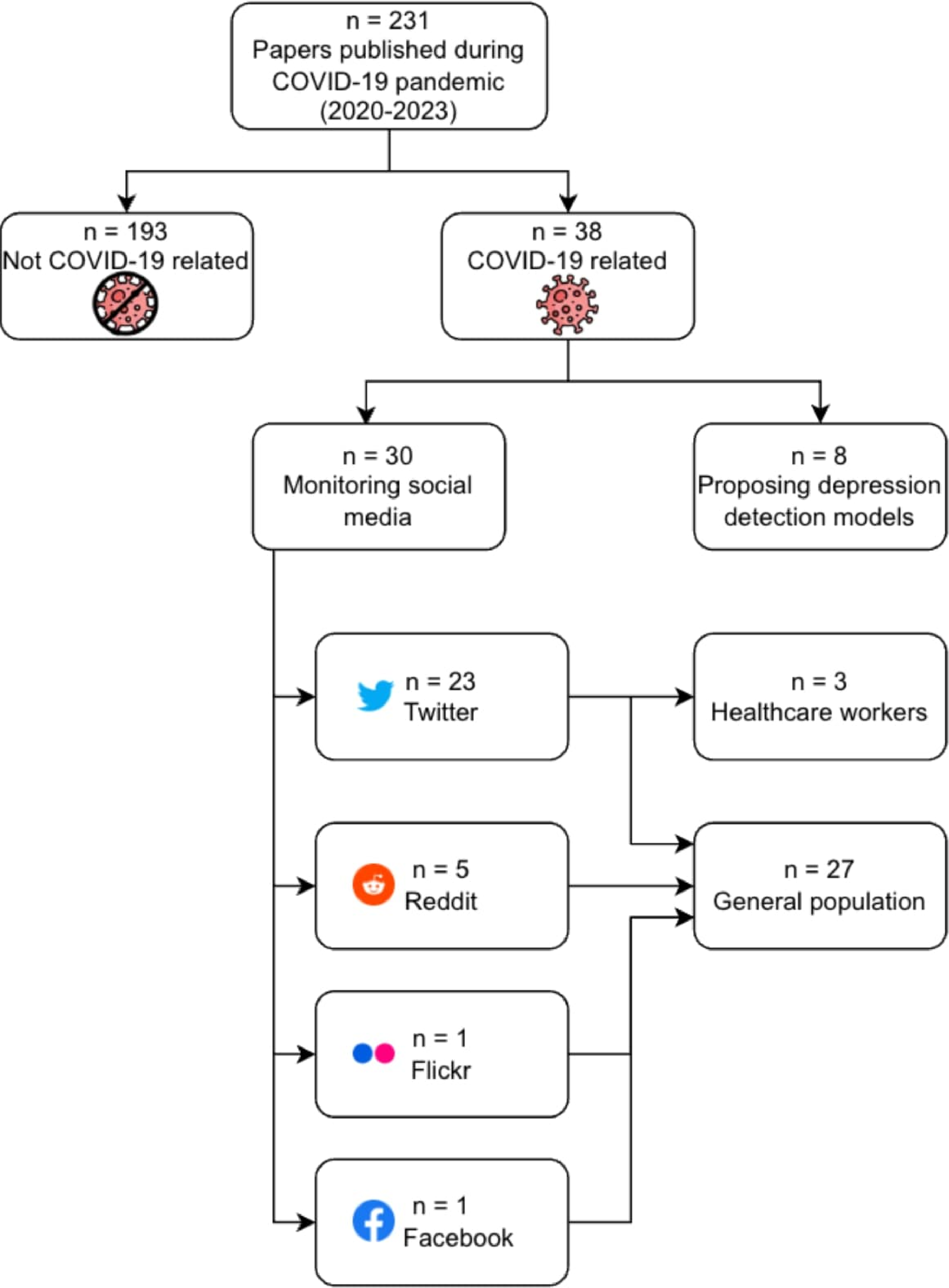}
	\caption{Taxonomy of papers published between 2020-2023. Only 38 out of the 231 papers addressed depression modeling in the context of the pandemic.}
	\label{covid-papers-figure}
\end{figure}

In Figure \ref{covid-papers-figure}, we propose a taxonomy of the 231 papers on depression detection published during the 2020-2023 period. Most articles (n = 193) tackling depression detection from social media data proposed general methods or datasets; only a small number of papers (n = 38) chose to work on assessing the impact of the global COVID-19 pandemic on the population's mental health. In this section, we present the COVID-19-related studies according to the classification from Figure \ref{covid-papers-figure}: studies monitoring social media posts and works proposing methods for depression detection in the context of the pandemic.

\subsection{Monitoring Social Media Streams During the COVID-19 Pandemic}
\label{sec:monitoring}

Most of the studies that use different social media platforms to monitor depression manifestations during the pandemic focus on data collected from Twitter. Other social media platforms of interest were Reddit, Flickr, and Facebook. For measuring the impact of the pandemic on mental health, most papers use keyword-based approaches for collecting depression-related content. Several studies use machine learning models to predict user- and post-level depression.

\vspace{2mm}
\noindent
\textbf{Twitter}\footnote{Twitter is currently known as X. Since all the research reported in this review was conducted before this name change, we use the term `Twitter' instead of `X' throughout the review}.~~
During the COVID-19 pandemic, Twitter was the leading social media platform for mental health surveillance. Two main approaches were chosen for real-time assessment of population-level mental health: keywords-based methods for retrieving and analyzing depression-related content and approaches using machine learning models for detecting the presence or absence of symptoms of mental disorders. Keyword-based analyses revealed a rise in depression-related tweets at the beginning of the pandemic \cite{bankston2022study, tommasel2022tracking}. Furthermore, using vocabulary-based methods, Cohrdes et al. (2021) \cite{cohrdes2021indications} extracted and monitored depression symptoms from Twitter data, such as depressed mood, sleep problems, fatigue, feelings of worthlessness, and diminished interest. With regard to automatic methods for social media monitoring, the studies published during the pandemic used various methods for estimating population-level depression from classical machine learning models \cite{biester2020quantifying,wolohan2020estimating,fine2020assessing,tabak2020temporal,zhou2021detecting} to state-of-the-art deep learning models \cite{kaseb2022analysis,li2020we,zhang2021monitoring}. 

\vspace{2mm}
\noindent
\textbf{Reddit~~}
Regarding Reddit, Biester et al. (2020) \cite{biester2020quantifying} analyzed the social media posts from three subreddits, \verb|r/Anxiety|, \verb|r/depression|, and \verb|r/SuicideWatch|, during January 2017 and May 2020, to measure the effect of the pandemic on posting patterns. Unlike previous studies on Twitter, which showed an increase in depression-related posts, the activity on \verb|r/depression| decreased at the beginning of the pandemic. By performing topic modeling, the authors revealed that the posts from home, family, and children topics increased significantly in the \verb|r/depression| subreddit. Similar to previous work, Biester et al. (2021) \cite{biester2021understanding} compared the activity in \verb|r/Anxiety|, \verb|r/depression|, and \verb|r/SuicideWatch| subreddits with control subreddits and showed that, while the activity in mental health-related subreddits decreased during the pandemic, the activity in control subreddits increased. Chen et al. (2022) \cite{chen2022covid} also showed that the sentiment in the \verb|r/depression| subreddit became more negative during the pandemic.

As opposed to the previous studies, which monitored specific subreddits, Wolohan (2020) \cite{wolohan2020estimating} monitored random users active in \verb|r/AskRedit| to avoid any biases from users' membership in specific depression-related subreddits. The author collected a novel dataset with users' Reddit posts from three distinct time spans: the first six months of 2018 and 2019 and the first four months of 2020. An LSTM model using fastText embeddings was trained on the Off-Topic Depression dataset \cite{wolohan-etal-2018-detecting} to predict user-level depression in each of the three periods of time analyzed. Compared to the periods from 2018 and 2019, the data showed an approx. 50\% increase in depression rate during the first four months of 2020 when the pandemic started. Bajaj et al. (2021) \cite{bajaj2021mental} monitored Reddit and showed that 6.4\% of users who did not have any depression-loaded posts before the pandemic began experiencing depression symptoms as detected by machine learning models during the pandemic.

\vspace{2mm}
\noindent
\textbf{Facebook~~}
Sukhwal and Kankanhalli (2022) \cite{sukhwal2022determining} studied posts from Facebook groups to determine the impact of containment policies on the mental well-being of the population. The analysis of topics discussed in these posts revealed that issues related to depression and suicide emerged after the lockdown, along with concerns related to the health and safety of frontline workers. Even after the containment policies were relaxed, the discussion of depression continued to increase, indicating that people were still struggling with the effects of the pandemic on their mental health.

\vspace{2mm}
\noindent
\textbf{Flickr~~}
In contrast to prior efforts that studied textual information, Fernández-Barrera et al. (2022) \cite{FernandezBarrera2022evaluating} analyzed the images with depression tags and their descriptions posted on Flickr. The pre-pandemic data was collected from January 2018 to December 2019, while the post-pandemic data was published between January 2020 and December 2021. The pre-pandemic posts contained tags associated with depression symptoms, such as ``sad", ``melancholic", and ``gloomy", while post-pandemic tags were related to feelings of loneliness and isolation. Furthermore, the visual analysis of the images revealed that there were more prevalent pre-pandemic images in the outdoor context than post-pandemic images with more indoor photos, which was expected due to the limited mobility imposed on the population.

\subsection{How Has the Pandemic Affected the Population?}
\label{sec:impact}

Studies for real-time mental health assessment in social media were performed to understand the impact of the COVID-19 pandemic on the population. In this subsection, we dive deeper into the papers from Subsection \ref{sec:monitoring}, to understand how healthcare workers and the general population from specific geographical regions were affected by the pandemic. In addition to the general population, healthcare workers were monitored since they were at risk for developing mental health problems due to increased workloads and direct involvement in treating those affected by the virus \cite{billings2021experiences}.

\subsubsection{General population}
Studies were carried out on population mainly from North America (US \cite{li2020modeling,saha2020psychosocial,fine2020assessing,zhang2021monitoring,kaseb2022analysis,zhang2022covid,li2022tracking,levanti2023depression},
Canada \cite{tshimula2022covid}), South America (Argentina \cite{tommasel2022tracking}). Fewer studies focused on the European continent \cite{tabak2020temporal,cohrdes2021indications,marshall2022using}, Australia \cite{zhou2021detecting} or Asia (Japan \cite{bashar2022tracking}, Singapore \cite{sukhwal2022determining}, Indian subcontinent \cite{gour2021depression}). There were other studies that weren't focused on a particular location \cite{li2020we,wolohan2020estimating,biester2020quantifying,davis2020quantifying,bajaj2021mental,el2021mental,biester2021understanding,nandy2021my,bankston2022study,FernandezBarrera2022evaluating,wu2023exploring}. These studies have shown an increase in the population rate of depression \cite{wolohan2020estimating} and in the depression symptoms reported by users in social media \cite{cohrdes2021indications,bankston2022study}. We are presenting below the findings from the studies that used monitoring methods for assessing the impact of the pandemic on the population from the geographical regions mentioned. Even if there was a high prevalence of depression due to the pandemic in the population from African countries, as shown by psychological studies \cite{bello2022prevalence}, there were no computational works focusing on this region, except for Wang et al. (2022) \cite{wang2022monitoring}, which monitored depression trends globally. Compared to other regions analyzed, Wang et al. (2022) \cite{wang2022monitoring} found fewer tweets geolocated in Africa and South America.

%%%%%%%%%%%%%%%%%%%%%%%
\vspace{2mm}
\noindent
\textbf{North America~~} In the \textbf{US}, the depression evolution was aligned with the national prevalence of COVID-19 cases \cite{li2020modeling,zhang2022covid,levanti2023depression}. Levanti et al. (2023) \cite{levanti2023depression} used classical machine learning models to detect depression in tweets from 7 major US cities with different response protocols and lockdown policies (Atlanta, Chicago, Houston, Los Angeles, Miami, New York, and Phoenix) between January and September 2019 and 2020. The depression evolution in these cities was aligned with the national prevalence of COVID-19 cases, rather than local trends or stay-at-home orders. Kaseb et al. (2022) \cite{kaseb2022analysis} showed that depressed tweets were more frequent than not-depressed ones and that the depression level increased from February 2020 to November 2021. Zhang et al. (2022) \cite{zhang2022covid} analyzed the mental health-related concerns on Twitter during the pandemic from different demographic groups and showed that Male, White, and aged between 30 and 49 years old users expressed most mental health concerns on Twitter. Moreover, topic analysis using Latent Dirichlet Allocation (LDA) revealed that most concerns shared on social media were related to social isolation, death toll, and politics and policies. Saha et al. (2020) \cite{saha2020psychosocial} used the Sparse Additive Generative Model (SAGE) \cite{eisenstein2011sparse} to identify language indicative of mental disorders and compared the 2020 period to 2019 to monitor for changes in mental health at the beginning of the COVID-19 pandemic in the US. The results showed a 21.32\% increase in anxiety, 19.73\% in suicidal ideation, 10.18\% increase in depression, and 3.76\% in stress. The expressions of social support also became more prevalent during the pandemic, with a 4.77\% increase in emotional support and 4.78\% in informational support. After the initial increase, the temporal analysis indicated a decline in mental health symptoms and support expressions.

Tshimula et al. (2022) \cite{tshimula2022covid} collected Twitter data during the stay-at-home period in \textbf{Canada} using COVID-19 related keywords and \#StayAtHome hashtag. LDA-based topic analysis on the Twitter data revealed that users were engaging more in depression-related topics during the stay-at-home period, with a rise in the second week of March, corresponding to the onset of the Canadian lockdown. Towards the end of the lockdown in May, Twitter users showed a decrease in participation in depressed-related topics.

%%%%%%%%%%%%%%%%%%%%%%%
\vspace{2mm}
\noindent
\textbf{South America~~}
Tommasel et al. (2022) \cite{tommasel2022tracking} used a lexicon-based approach to analyze Twitter data from \textbf{Argentina} between March and August 2020 to monitor the changes in mental health during the lockdown. The rise in depression-related posts began with the confirmation of the first COVID-19 case in Argentina and the establishment of the lockdown. Moreover, a high prevalence of depression-related posts was observed when the lockdown extension was announced, with Twitter users expressing health and economic concerns.

%%%%%%%%%%%%%%%%%%%%%%%
\vspace{2mm}
\noindent
\textbf{Europe~~}
Tabak and Purver (2020) \cite{tabak2020temporal} analyzed the temporal depression dynamics in relation to the lockdowns from \textbf{France}, \textbf{Germany}, \textbf{Italy}, \textbf{Spain} and the \textbf{United Kingdom}. One Bi-directional LSTM model with self-attention was trained on Twitter data from 2015 to October 2019 for each of the five locations of interest to predict depression. The models were further used to predict depression for each day from December 1, 2019, until May 15, 2020, using all the posts from a given day. The authors compared the depression rate before and during the lockdown for each country. At the beginning of the lockdown, there was an increase in depressed tweets from French, Italian, German, and Spanish users. Regarding the UK, there had been an increase in depression before the national lockdown, which afterward plateaued when the lockdown began. The easing of the lockdown restrictions corresponded to an improvement in mental state for the French and Spanish populations, with the depression rate in Italy and Germany further increasing. In contrast, the UK population was agnostic to the change. Another study on the \textbf{UK} population between July 23, 2020, and January 6, 2021, had shown that the amount of depression-related tweets increased before the second national lockdown began, followed by a decrease and another increase with the enforcement of the third national lockdown \cite{marshall2022using}.

In a study by Cohrdes et al. (2021) \cite{cohrdes2021indications}, depression symptoms on Twitter were compared to survey-based self-reports using PHQ-8 between January and July 2020 in \textbf{Germany}, before, during, and after the first lockdown. Posts with depression symptoms were extracted from Twitter data using a vocabulary-based method. The study found a moderate to high correlation between depression symptoms measured through survey and Twitter data for most of the symptoms of interest, such as depressed mood, sleep problems, fatigue, and feelings of worthlessness, except for diminished interest. Unlike previous studies that showed high depressive symptoms during the pandemic and lockdowns, the German population demonstrated a temporary decrease in some symptoms during the lockdown. The study highlights the potential of Twitter data to monitor users' mental states online and provide early intervention measures.

%%%%%%%%%%%%%%%%%%%%%%%
\vspace{2mm}
\noindent
\textbf{Asia~~} Bashar (2022) \cite{bashar2022tracking} compared the depression levels of people from \textbf{Japan} before and during the COVID-19 pandemic. A logistic regression model was used for classifying tweets into depressive and non-depressive categories. The author showed a higher prevalence of tweets categorized as depressive during the pandemic, with 50.37\% tweets classified as depressive, out of the total COVID-19-related tweets. The depression level of Japanese Twitter users increased considerably after the peak of COVID-19 deaths in Japan. 

Gour et al. (2021) \cite{gour2021depression} performed real-time depression analysis on Twitter streams in countries from the Indian subcontinent (\textbf{India, Bhutan, Pakistan, Nepal}). The authors monitored Twitter users' emotions and sentiments using lexicon and machine learning approaches (Decision Trees, Naive Bayes) to understand the depression dynamics of people from the countries above. Pakistan had the largest number of negative sentiment tweets, while also having high levels of posts with fear and anger.

Sukhwal and Kankanhalli (2022) \cite{sukhwal2022determining} investigated Facebook posts from \textbf{Singapore} to measure the influence of containment policies on the population's mental health. Topic analysis showed the emergence of depression and suicide topics after the lockdown was announced in Singapore. With the relaxation of containment policies, the topic of depression became even more prevalent, indicating that people were still suffering due to the pandemic.

%%%%%%%%%%%%%%%%%%%%%%%
\vspace{2mm}
\noindent
\textbf{Australia~~} Zhou et al. (2021) \cite{zhou2021detecting} studied the depression dynamics in the population of New South Wales, \textbf{Australia}, using Twitter data from January to May 2020. A logistic regression model with TF-IDF, emotional, topic, and domain-specific features was used to detect depression in Twitter data posted during the pandemic. The results showed that the community depression level increased proportional to the number of COVID-19 cases and then decreased after the end of March when the COVID-19 cases peaked.

As opposed to previous studies, which monitor tweets geolocated in a specific region, Wang et al. (2022) \cite{wang2022monitoring} collect and analyze depression and COVID-19-related posts from different geographical areas: North America, South America, Europe, Asia, Africa, and Oceania. The study aimed to uncover the depression trends before and after the release of the COVID-19 vaccine. A keyword-based analysis revealed that different themes were present in the depression-related posts during the pandemic, but after the release of the vaccine, a common theme of returning to normal social activities was prevalent.

\subsubsection{Healthcare workers}

Healthcare workers were more susceptible than the general population to mental illness during the pandemic, as they were on the frontline taking care of patients during the emergency situation \cite{spoorthy2020mental}. All studies on the mental well-being of healthcare workers were using Twitter data geolocated in the \textbf{US}. Fine et al. (2020) \cite{fine2020assessing} extracted and analyzed real-time mental health information from the social media platforms of healthcare providers. The authors collected and compared a sample of posts from healthcare professionals to posts from a control group comprised of people not working in the health sector. Logistic regression models were trained on n-gram features to predict if a post was written by a person suffering from depression, anxiety, or at risk of suicide. The study showed that healthcare professionals experienced higher levels of depression, anxiety, and suicide risk than the control group before the lockdown began. During the lockdown, the general population showed higher anxiety levels than healthcare providers. However, healthcare workers showed higher levels of suicide risk at the beginning of the lockdown. The changes in depression levels of both groups did not differ significantly during the period analyzed. 

Another study monitored Twitter data from February 2020 to April 2022 to measure the impact of lockdown policies on the population's mental health \cite{li2022tracking}. The authors compared mental health and COVID-19-related posts of healthcare workers to the ones from the general population. A lexicon-based approach was used to categorize mental health-related posts into four categories: anxiety, depression, addiction, and insomnia. All the mental health problems studied were more prevalent in the posts of healthcare professionals than in the general population. 

Agarwal et al. (2023) \cite{agarwal2023investigating} analyzed Twitter posts from emergency physicians and resident physicians in the US before and during the pandemic using different psychological measures for anxiety, depression, loneliness, and anger. The authors showed an increase in posts related to the four psychological measures during the pandemic. Furthermore, topic modeling analysis revealed that, during the pandemic, physicians expressed more concerns regarding pandemic response, healthy behaviors during the pandemic, vaccines, unstable housing, and providing emotional support.

\subsection{Depression Detection in the Context of the COVID-19 Pandemic}
\label{sec:detection-covid}

Few papers have proposed novel methods for depression detection in the COVID-19 context. We expand on some of the works in this section.

Given the increase in depression levels caused by the pandemic, Wu et al. (2023) \cite{wu2023exploring} proposed a method for early detection of depression for COVID-19 patients. One of the contributions of the paper was the DepCOV dataset. The data collection consisted of retrieving users with COVID-19 infection mentions from Twitter. These users were split into two groups: the control group, which did not mention depression before or after the COVID-19 diagnosis, and the group with depression signals, which had at least three depression-related posts after the infection, but none before infection. The authors proposed a novel transformer-based method, Mood2Content, to detect depression in COVID-19 patients using content and mood representations as early as possible. The model trained on the DepCOV dataset achieved a performance of 0.9317 AUROC in predicting COVID-19 patients at risk of depression two weeks before their first depression mention. Topic analysis on the social media content with depression mentions from COVID-19  patients revealed that concerns related to the pandemic, such as lockdown, government, policies, and treatment were predominant. 

Cha et al. (2022) \cite{cha2022lexicon} experimented with different architectures (BiLSTM, CNN, BERT-based) for the early detection of depression in English, Korean, and Japanese languages during the pandemic. Twitter data was collected from January 2019 to March 2021 to train the models. For each of the three languages, a lexicon-based approach was used to label posts from users in one of the two categories: depressed and non-depressed. The proposed models were trained on the labeled data to predict depression at the post-level and obtained good results of over 98\% F1-score. The authors then tested the models' generalizability by applying them to posts from Everytime, a community platform for Korean undergraduate students, during and before the pandemic. While the performance of the CNN and BiLSTM-based architectures degraded when predicting on the Everytime data, the BERT-based model had the greatest performance for this use case, with an F1-score of 64\%.

Zogan et al. (2023) \cite{zogan2023hierarchical} proposed a novel method based on Hierarchical Convolutional Neural Networks for depression detection during the pandemic. The proposed method considered the user's history, and the attention mechanism learned the most important words and posts from the timeline. The model was trained on the dataset from Shen et al. (2017) \cite{ijcai2017p536} and achieved an 86.9\% F1-score. The model was further used to identify users with depression using data from September 2019 to April 2020. The number of users predicted as depressed peaked in March when COVID-19  was declared a pandemic. LSTM and CNN-based neural network architectures were also used for depression detection during the COVID-19 pandemic \cite{anbalagan2022detecting,meena2022depression,al2023hybrid}.

Harrigian and Dredze (2022) \cite{harrigian2022problem} studied the semantic shift problem in longitudinal social media data monitoring during the COVID-19 pandemic. The authors used various depression datasets sampled from different social media platforms and time spans to measure semantic shift's influence on models for depression prevalence estimation. The classical approach for monitoring social media for depression is to train a statistical classifier on existing datasets and use the model to identify language associated with depression on new data collected from online platforms. Semantic shift, which occurs in longitudinal posts, can affect the performance of the models due to changes in the terminology for representing concepts or in the language distribution \cite{harrigian2022problem}. The authors showed that semantic shift increased with the time gap between training and evaluation data collection. When training on historical data, small changes in the vocabulary can lead to large deviations for downstream monitoring tasks, even if these changes do not affect the performance of the models on the historical data. Furthermore, the authors used semantic shift measurements to curate vocabularies with semantic stability, improving statistical classifiers' predictive generalization. When comparing the most common social media platforms for COVID-19 monitoring, using vocabularies with semantic stability was more useful in the case of Twitter than on Reddit, due to the particularities of each social media platform. On Twitter, conversations focus on current events, while Reddit fosters discussion on specific interests (in subreddits) evolving over extended periods.

\begin{table*}[h!]
    \centering
    \caption{List of available datasets for depression modeling using data posted on online platforms during the pandemic.}
    \resizebox{1\textwidth}{!}{
    \begin{tabular}{p{4cm}|p{1.2cm}|p{1.1cm}|p{3.8cm}|p{2.6cm}|p{1.9cm}}
        \textbf{Dataset}& \textbf{Platform}& \textbf{Language}&\textbf{Annotation Procedure} & \textbf{Dataset Size} & \textbf{Availability}\\
        \hline
        Zhang et al. (2021) \cite{zhang2021monitoring}  & Twitter & English & Self-disclosure & 5K users & UNK \\
        Cohrdes at al. (2021) \cite{cohrdes2021indications}  & Twitter & German & PHQ-8 symptoms-related terms & 88K posts & UNK\\
        Davis et al. (2022) \cite{articlesdht}  & Reddit & English & Subreddit participation & 81K users & UNK \\
        Fernández-Barrera et al. (2022) \cite{inproceedingsds}  & Flickr & English &  Depression tags & 14.5K posts & UNK\\
        Cha et al. (2022) \cite{cha2022lexicon}  & Twitter, Everytime & Korean, English, Japanese & Lexicon-based annotation & 26M posts (Twitter) 22K posts (Everytime) & FREE (sample)/ AUTH (full) \\
        Zogan et al. (2023) \cite{zogan2023hierarchical}  & Twitter & English & Self-disclosure & 1.4K users & UNK\\
        Wu et al. (2023) \cite{wu2023exploring} & Twitter & English& Self-disclosure & 10K users & DUA \\

    \end{tabular}
    }
    \label{tab:datasets}
\end{table*}

\section{Datasets for depression detection during COVID-19 pandemic}
\label{sec:datasets}
This section presents the particularities of the datasets from works on depression modeling during the COVID-19 pandemic\footnote{The list of datasets can also be consulted at:
\href{https://github.com/bucuram/depression-datasets-nlp}{https://github.com/bucuram/depression-datasets-nlp}}. Of the 38 COVID-19-related papers published during the pandemic, only 7 of them collect and annotate data from online platforms. Most research papers from Section \ref{sec:covid-19} collect social media data for monitoring the impact of COVID-19 on the population's mental health, without providing any annotations. We focus on the datasets that have annotations \cite{zhang2021monitoring,cohrdes2021indications,articlesdht,inproceedingsds,cha2022lexicon,zogan2023hierarchical,wu2023exploring}. For each dataset, we present in Table \ref{tab:datasets} information about the platform used for data collection, the language of the data, the annotation procedure, the size of the dataset and its availability.

\noindent \textbf{Availability~~} Given the sensitive information comprised in the datasets for depression modeling, their availability varies.  Most authors of the datasets listed in Table \ref{tab:datasets} do not specify the availability of their data; therefore, this information is categorized as unknown (UNK). In one case, the authors provided a sample of their data (FREE), while access to the remaining data is available upon request (AUTH). Only one dataset, from Wu et al. (2023) \cite{wu2023exploring}, requires a signed data agreement (DUA) for access.

\vspace{2mm}
\noindent \textbf{Platform~~} Twitter and Reddit are the most used online platforms for collecting data due to the availability and easy access to dedicated APIs. However, due to relatively recent changes in terms of service and API rate limits for Twitter\footnote{\href{https://developer.twitter.com/en/docs/twitter-api/rate-limits}{https://developer.twitter.com/en/docs/twitter-api/rate-limits}} and Reddit\footnote{\href{https://support.reddithelp.com/hc/en-us/articles/16160319875092-Reddit-Data-API-Wiki}{https://support.reddithelp.com/hc/en-us/articles/16160319875092-Reddit-Data-API-Wiki}}, data acquisition on these online platforms was made more difficult. Moreover, these recent changes might hinder the reproduction of datasets where the authors make available only the Twitter or Reddit IDs and not the raw texts. Furthermore, these changes make collecting new data challenging and time-consuming. For the pandemic-related datasets, the data was primarily collected from Twitter, while a few studies also included data from Reddit, Flickr, and Everytime.

\vspace{2mm}
\noindent \textbf{Language~~} Most of the data is in English, because it is the most predominant language on social media platforms. However, some of the data in Table \ref{tab:datasets} was in German, Korean or Japanese.

\vspace{2mm}
\noindent \textbf{Annotation procedure~~} Most of the datasets used in the 38 COVID-19-related papers were collected with the scope of monitoring and did not undergo any labeling process. From the datasets with annotated data in Table \ref{tab:datasets}, three of them are based on the self-disclosure of depression diagnoses. This annotation process consists of automatically labeling users binary by the presence or absence of mentions of depression diagnosis in their posts. Using this approach, large datasets with hundreds of thousands of users can be collected. However, even if annotators manually inspect the posts with self-disclosure, there is no way to verify if the disclosure is authentic and if the user is telling the truth. Furthermore, regarding the control group, in which users do not have any mentions of a mental health diagnosis, their real mental health status is unknown; we cannot imply that they do not have any mental disorders only by the absence of self-disclosure. It is important to note that relying on self-reported diagnoses to collect data about mental health can lead to self-selection bias \cite{amir2019mental}. This means that the collected data may only represent individuals who openly discuss their mental health diagnosis and may not accurately reflect the entire population of people diagnosed with mental disorders. Other methods used to annotate the pandemic data were based on lexicons, tags, symptoms-related terms or subreddit participation.

\section{Discussion}
\label{sec:discussion}
The COVID-19 pandemic had a significant impact on people's mental well-being. Social media platforms turned out to be an ideal medium for measuring the impact of the pandemic on the population's mental health, with results comparable to survey-based approaches used in psychology research \cite{cohrdes2021indications}. However, in the context of the COVID-19 pandemic, most research published between 2020-2023 uses similar methods for estimating population depression prevalence, such as training machine learning classifiers on previously available datasets with data from before the pandemic. These classifiers are then used to predict the depression status of new data posted during the pandemic. As presented in Subsection \ref{sec:detection-covid}, training the models on available depression datasets from different time periods may affect their generalizability on new data, due to semantic shift, leading to misleading conclusions (e.g., an increase in depression levels, instead of decrease) \cite{harrigian2022problem}. One way to address this problem is to train and evaluate the model on data from the same temporal distribution or to account for the semantic shift by performing semantically informed feature selection. Another technique to avoid performance degradation in word embeddings or pre-trained models is to measure semantic shift stability and retrain the model accordingly \cite{ishihara2022semantic}.

Social media platforms provide a vast amount of information that can be used to identify and track mental health disorders, such as depression. Machine learning algorithms analyze patterns in language use and detect signals of mental health conditions. In recent years, there has been a shift from binary classification models for predicting depression status to explainable algorithms that make predictions based on depressive symptoms using social media data. Researchers leverage social media data annotated for symptoms measured through psychological assessment tools (PHQ-9, BDI) or described in DSM-V to develop methods for explainable mental disorders detection. For instance, Zhang et al. 
(2022) \cite{zhang-etal-2022-symptom} developed an automatic symptom-assisted method for mental disorders detection that provides the corresponding mental health symptoms that lead to the specific decision for each prediction. Furthermore, automated depression detection models trained to detect symptoms of mental disorders or contained by the presence of symptoms have shown greater generalizability than classical binary models \cite{nguyen2022improving,perez2023bdi}. To interpret the predictions of automatic models, other works use the attention weights \cite{uban2021emotion,ghosh2023attention}, SHapley Additive exPlanation (SHAP) \cite{lundberg2017unified,bucur2021early} or Local Interpretable Model-agnostic Explanations (LIME) \cite{ribeiro2016should,ADARSH2023103168}. 

Besides explaining or interpreting the prediction of automatic depression detection models, another critical aspect is model calibration. Calibration refers to the capability of the model's output to accurately reflect the chance of a correct prediction \cite{desai2020calibration,ulmer2022exploring}. For example, in the case of depression, a model that predicts the depression class with 70\% confidence is considered well-calibrated if the event occurs in 70\% of the cases when that confidence level is assigned. Desai and Durrett (2020) \cite{desai2020calibration} examine the calibration of two pre-trained transformer-based models, BERT \cite{kenton2019bert} and RoBERTa \cite{liu2019roberta}, which are often employed in depression modeling. The authors compare the models on different downstream tasks and show that both models achieve low expected calibration errors. However, RoBERTa has a higher performance and lower in- and out-of-domain calibration errors. There is only one work developing depression detection models that accounts for model calibration. Ilias et al. (2023) \cite{ilias2023calibration} inject extra-linguistic information to MentalBERT \cite{ji2022mentalbert}, a transformer-based model pre-trained on mental health data, and BERT. Furthermore, the authors employed label smoothing, which resulted in better-performing and calibrated models.

Transparency is a crucial aspect of documenting decision-making and enhancing research practice in current ethics practices \cite{ajmani2023systematic}. There has been a growing dialogue surrounding the use of social media data for research purposes. Individuals who generate online content, such as tweets and blog posts, are often unaware that their contributions may be used in research endeavors, particularly concerning mental health studies \cite{birnbaum2017collaborative}. However, mining the data for social media research is increasingly prevalent. Scholars have advocated for broadening the scope of social media ethics to address specific practices, such as user privacy, data attribution, and informed consent \cite{ayers2018don}. While AI and ML present great potential for identifying mental health disorders, researchers must uphold ethical principles of privacy, confidentiality, and respect for the well-being of the participants by being transparent about their methodologies and taking steps to mitigate biases. This responsible approach allows for a more comprehensive understanding of mental health conditions while safeguarding the rights and dignity of the individuals involved in the research process. Data anonymization and de-identification are necessary to protect the confidentiality of individuals. This involves removing any personally identifiable information to minimize the risk of re-identification, which could expose sensitive mental health information \cite{benton2017ethical}. However, because not all studies on mental disorders modeling disclose if they have ethical approval from the Institutional Review Board (IRB) for their research, there is no way to ensure that the data were processed by taking into consideration the privacy of the users. Ajmani et al. (2023) \cite{ajmani2023systematic} reviewed ethics disclosures in articles that employ machine learning models for inferring the mental health status of social media users and showed that the trend in the rate of ethics disclosures is slowly increasing. Moreover, ethical considerations from research articles are more often data-driven practices than human-centered ones. Data-centered practices comprise information about subjects' inclusion and exclusion criteria, how the data is stored or shared, and whether the data was de-identified or modified when presented in the form of quotes or examples in research papers. On the other hand, a human-centered approach to ethics disclosures consists of information on whether the data are public or private, whether ethical approval from IRB was granted, how the researchers interacted with the participants, if informed consent was received and compensations were offered, and if the findings of the study will be shared with the subjects. Even though mental disorders modeling from social media data is a human-centered task, human-centered ethical disclosures are not prioritized \cite{ajmani2023systematic}. To improve ethical disclosures and increase transparency in ML-based mental health research, Ajmani et al. (2023) \cite{ajmani2023systematic} proposed several solutions. These include interdisciplinary discussions on ethical practices in dedicated panels, dedicated space for ethical considerations in research papers, and employing context documents for models \cite{mitchell2019model} and datasets \cite{boyd2021datasheets}.

The exploration of mental health in social media data is often limited to English data, without considering individuals' demographic information. Some of the popular datasets, such as CLPsych 2015 \cite{coppersmith2015clpsych} or MULTITASK \cite{benton2017multitask}, have been analyzed by Aguirre et al. (2021) \cite{aguirre2021gender} and found to be demographically unrepresentative, with more female users and an over-representation of White users. This demographic imbalance has resulted in poorer performance of models for underrepresented groups. To improve the model's performance for underrepresented demographic groups, Aguirre et al. (2021) \cite{aguirre2021gender} reported that increasing the data size and balancing it can be beneficial. Moreover, previous research has shown that mental illnesses have different manifestations in social media language, depending on factors like age, gender, or race \cite{de2017gender,loveys-etal-2018-cross,amir2019mental,aguirre2021gender,aguirre2021qualitative}. Even if pre-trained transformer-based models represent the state-of-the-art in mental health research, they can capture gender-related social stigma in mental health. Lin et al. (2022) \cite{lin2022gendered} analyzed RoBERTa and MentalRoBERTa and showed that both models are more likely to predict female subjects in sentences about mental disorders. However, all these studies have been conducted on English-speaking subjects, and there has been no exploration to determine whether the manifestations change based on the subjects' mother tongue. With the availability of datasets and dedicated shared tasks in languages other than English, such as MentalRiskES \cite{marmol2023overview}, exploring demographic differences in the manifestations of depression on social media represents a new avenue for future research.

\section{Conclusion}
\label{sec:conclusion}
This paper presented the most comprehensive and up-to-date review of NLP approaches developed for modeling depression in social media. We make an overview of papers published in international conferences and journals between 2020 and 2023. We focus on the impact of the COVID-19 pandemic on mental health and depression detection, providing the reader with a post-COVID-19 pandemic outlook on the topic. 

We identified multiple trends and open challenges, which we summarize as follows:

\begin{itemize}
    \item We observed that modeling depression using social media continues to be a relevant problem and that the interest in it continues to grow. Our review shows that the number of papers published every year has been on an uphill trend since the beginning of pandemic. We have seen a spike in 2023, partially motivated by COVID-19. We believe that the unprecedented public health emergency caused by the pandemic and the subsequent increase in the prevalence of mental health disorders are among the reasons for the increased interest in the topic. Furthermore, recent advances in AI/NLP models, such as pre-trained LLMs and neural networks, may also have helped fuel further interest in this topic. We believe that the impact of the pandemic and technological advances are likely to continue motivating research in depression detection in the next few years.
    \item User-generated content continues to be a relevant source of information for modeling depression. Models of depression detection have shown to be reliable, and some studies \cite{articleeich,kelly2020blinded,cohrdes2021indications} have evidenced correlation with survey-based methods or patient's medical history. 
    \item The availability of suitable data is an issue. Social media platforms such as Twitter and Reddit have recently changed their data access and distribution policies, opting for more restrictive policies. This will impact both new and existing datasets collected from these platforms. Researchers will have to find alternative ways to leverage user-generated content for mental health analysis. This includes specialized mental health forums and/or other social platforms that are less restrictive with regard to data access and distribution.
\end{itemize}

\noindent As discussed in this review, modeling depression in social media continues to receive sustained attention from the research community. We see the following research directions being explored in the near future.

\begin{itemize}
\item We expect the use of generative AI and LLMs in particular (e.g., BLOOM \cite{workshop2022bloom}, GPT-3.5 \cite{openai2022chatgpt}, LLaMA \cite{touvron2023llama}) to increase in future research \cite{amin38will}. LLMs have been used for providing explanations for predictions \cite{yang2023evaluations,qin2023read} or generating synthetic depression data aiming to circumvent the restrictions on data collection and distribution imposed by social media platforms \cite{bucur2023utilizing}.
\item Multi-modal models have become increasingly popular in recent years as neural networks have been shown to jointly model speech, text, and images with improved accuracy. Such models have been used in various scenarios, including modeling mental health states and conditions from social media data \cite{bucur2023matter,yadav2023towards,gimeno2024reading}. We anticipate more multi-modal datasets to become available, which, combined with recent advances in multi-modal LLMs such as the GPT 4.0 \cite{openai2023gpt} and LLaVA \cite{liu2023visual}, will contribute to fueling research on multi-modal models of depression.
\item Various high-performing pre-trained multi-lingual models (e.g., mBERT \cite{pires2019multilingual}, XLM-R \cite{conneau2020unsupervised}), each supporting hundreds of languages have been made available in the past few years. However, social media datasets collected and annotated for depression contain data in English and a couple of other high-resource languages, making it challenging to develop multi-lingual systems for depression detection. Multi-lingual datasets are still not widely available for research purposes, hindering progress on research on languages other than English. We expect more datasets in languages other than English to be released in the near future, motivating research on this topic in low-resource languages. 
\end{itemize}

\section*{References}
\bibliographystyle{IEEEtran}
\bibliography{refs}

\begin{thebibliography}{100}
\providecommand{\url}[1]{#1}
\csname url@samestyle\endcsname
\providecommand{\newblock}{\relax}
\providecommand{\bibinfo}[2]{#2}
\providecommand{\BIBentrySTDinterwordspacing}{\spaceskip=0pt\relax}
\providecommand{\BIBentryALTinterwordstretchfactor}{4}
\providecommand{\BIBentryALTinterwordspacing}{\spaceskip=\fontdimen2\font plus
\BIBentryALTinterwordstretchfactor\fontdimen3\font minus \fontdimen4\font\relax}
\providecommand{\BIBforeignlanguage}[2]{{%
\expandafter\ifx\csname l@#1\endcsname\relax
\typeout{** WARNING: IEEEtran.bst: No hyphenation pattern has been}%
\typeout{** loaded for the language `#1'. Using the pattern for}%
\typeout{** the default language instead.}%
\else
\language=\csname l@#1\endcsname
\fi
#2}}
\providecommand{\BIBdecl}{\relax}
\BIBdecl

\bibitem{davey2019early}
C.~G. Davey and P.~D. McGorry, ``Early intervention for depression in young people: a blind spot in mental health care,'' \emph{The Lancet Psychiatry}, vol.~6, no.~3, pp. 267--272, 2019.

\bibitem{read2022history}
H.~Read and B.~A. Kohrt, ``The history of coordinated specialty care for early intervention in psychosis in the united states: A review of effectiveness, implementation, and fidelity,'' \emph{Community Mental Health Journal}, pp. 1--12, 2022.

\bibitem{chancellor2020methods}
S.~Chancellor and M.~De~Choudhury, ``Methods in predictive techniques for mental health status on social media: a critical review,'' \emph{NPJ digital medicine}, vol.~3, no.~1, pp. 1--11, 2020.

\bibitem{wolohan2020estimating}
J.~Wolohan, ``Estimating the effect of covid-19 on mental health: Linguistic indicators of depression during a global pandemic,'' in \emph{Proc. of NLP-COVID19 Workshop, EMNLP}, 2020.

\bibitem{kaseb2022analysis}
A.~Kaseb, O.~Galal, and D.~Elreedy, ``Analysis on tweets towards covid-19 pandemic: An application of text-based depression detection,'' in \emph{Proceedings od NILES}.\hskip 1em plus 0.5em minus 0.4em\relax IEEE, 2022, pp. 131--136.

\bibitem{lenharo2023declares}
M.~Lenharo, ``Who declares end to covid-19's emergency phase,'' \emph{Nature}, vol. 882, no. 10.1038, 2023.

\bibitem{calvo2017natural}
R.~A. Calvo, D.~N. Milne, M.~S. Hussain, and H.~Christensen, ``Natural language processing in mental health applications using non-clinical texts,'' \emph{Natural Language Engineering}, vol.~23, no.~5, pp. 649--685, 2017.

\bibitem{skaik2020using}
R.~Skaik and D.~Inkpen, ``Using social media for mental health surveillance: a review,'' \emph{ACM Computing Surveys}, vol.~53, no.~6, pp. 1--31, 2020.

\bibitem{rissola2021survey}
E.~A. R{\'\i}ssola, D.~E. Losada, and F.~Crestani, ``A survey of computational methods for online mental state assessment on social media,'' \emph{ACM Transactions on Computing for Healthcare}, vol.~2, no.~2, pp. 1--31, 2021.

\bibitem{zhang2022natural}
T.~Zhang, A.~M. Schoene, S.~Ji, and S.~Ananiadou, ``Natural language processing applied to mental illness detection: a narrative review,'' \emph{NPJ digital medicine}, vol.~5, no.~1, p.~46, 2022.

\bibitem{dhelim2023detecting}
S.~Dhelim, L.~Chen, S.~K. Das, H.~Ning, C.~Nugent, G.~Leavey, D.~Pesch, E.~Bantry-White, and D.~Burns, ``Detecting mental distresses using social behavior analysis in the context of covid-19: A survey,'' \emph{ACM Computing Surveys}, 2023.

\bibitem{losada2016test}
D.~E. Losada and F.~Crestani, ``A test collection for research on depression and language use,'' in \emph{Proc. of CLEF}.\hskip 1em plus 0.5em minus 0.4em\relax Springer, 2016, pp. 28--39.

\bibitem{coppersmith2015clpsych}
G.~Coppersmith, M.~Dredze, C.~Harman, K.~Hollingshead, and M.~Mitchell, ``Clpsych 2015 shared task: Depression and ptsd on twitter,'' in \emph{Proc. of CLPsych Workshop, NAACL}, 2015, pp. 31--39.

\bibitem{zirikly2019clpsych}
A.~Zirikly, P.~Resnik, O.~Uzuner, and K.~Hollingshead, ``Clpsych 2019 shared task: Predicting the degree of suicide risk in reddit posts,'' in \emph{Proc. of CLPsych Workshop, NAACL}, 2019, pp. 24--33.

\bibitem{lynn2018clpsych}
V.~Lynn, A.~Goodman, K.~Niederhoffer, K.~Loveys, P.~Resnik, and H.~A. Schwartz, ``Clpsych 2018 shared task: Predicting current and future psychological health from childhood essays,'' in \emph{Proc. of CLPsych Workshop, NAACL}, 2018, pp. 37--46.

\bibitem{kayalvizhi2022findings}
S.~Kayalvizhi, T.~Durairaj, B.~R. Chakravarthi \emph{et~al.}, ``Findings of the shared task on detecting signs of depression from social media,'' in \emph{Proc. of LTEDI Workshop}, 2022, pp. 331--338.

\bibitem{s-etal-2023-overview-shared}
K.~S, T.~D., B.~R. Chakravarthi, J.~M. C, K.~S~V, and P.~A. Rahood, ``Overview of the shared task on detecting signs of depression from social media text,'' in \emph{Proc. of LTEDI Workshop}, B.~R. Chakravarthi, B.~Bharathi, J.~Griffith, K.~Bali, and P.~Buitelaar, Eds.\hskip 1em plus 0.5em minus 0.4em\relax Varna, Bulgaria: INCOMA Ltd., Shoumen, Bulgaria, Sep. 2023, pp. 25--30.

\bibitem{marmol2023overview}
A.~M. M{\'a}rmol-Romero, A.~Moreno-Mu{\~n}oz, F.~M. Plaza-del Arco, M.~D. Molina-Gonz{\'a}lez, M.~T. Mart{\'\i}n-Valdivia, L.~A. Ure{\~n}a-L{\'o}pez, and A.~Montejo-Ra{\'e}z, ``Overview of mentalriskes at iberlef 2023: Early detection of mental disorders risk in spanish,'' \emph{Procesamiento del Lenguaje Natural}, vol.~71, pp. 329--350, 2023.

\bibitem{cohan2018smhd}
A.~Cohan, B.~Desmet, A.~Yates, L.~Soldaini, S.~MacAvaney, and N.~Goharian, ``Smhd: a large-scale resource for exploring online language usage for multiple mental health conditions,'' \emph{arXiv preprint arXiv:1806.05258}, 2018.

\bibitem{Eichstaedt11203}
J.~C. Eichstaedt, R.~J. Smith, R.~M. Merchant, L.~H. Ungar, P.~Crutchley, D.~Preo{\c t}iuc-Pietro, D.~A. Asch, and H.~A. Schwartz, ``Facebook language predicts depression in medical records,'' \emph{PNAS}, vol. 115, no.~44, pp. 11\,203--11\,208, 2018.

\bibitem{jamil-etal-2017-monitoring}
Z.~Jamil, D.~Inkpen, P.~Buddhitha, and K.~White, ``Monitoring tweets for depression to detect at-risk users,'' in \emph{Proc. of CLPsych Workshop, NAACL}.\hskip 1em plus 0.5em minus 0.4em\relax Vancouver, BC: ACL, Aug. 2017, pp. 32--40.

\bibitem{Yazdavar2020MultimodalMH}
A.~H. Yazdavar, M.~S. Mahdavinejad, G.~Bajaj, W.~L. Romine, A.~Sheth, A.~H. Monadjemi, K.~Thirunarayan, J.~M. Meddar, A.~C. Myers, J.~Pathak, and P.~Hitzler, ``Multimodal mental health analysis in social media,'' \emph{PLoS ONE}, vol.~15, 2020.

\bibitem{ijcai2017p536}
G.~Shen, J.~Jia, L.~Nie, F.~Feng, C.~Zhang, T.~Hu, T.-S. Chua, and W.~Zhu, ``Depression detection via harvesting social media: A multimodal dictionary learning solution,'' in \emph{Proc. of IJCAI}, 2017, pp. 3838--3844.

\bibitem{ALSAGRI_2020}
H.~S. Alsagri and M.~Ykhlef, ``Machine learning-based approach for depression detection in twitter using content and activity features,'' \emph{IEICE}, vol. E103.D, no.~8, pp. 1825--1832, aug 2020.

\bibitem{article23}
Z.~Peng, Q.~Hu, and J.~Dang, ``Multi-kernel svm based depression recognition using social media data,'' \emph{Int. J. of ML and Cybernetics}, vol.~10, 01 2019.

\bibitem{bucur2021psychologically}
A.-M. Bucur, I.~R. Podin{\u{a}}, and L.~P. Dinu, ``A psychologically informed part-of-speech analysis of depression in social media,'' in \emph{Proc. of RANLP}, 2021, pp. 199--207.

\bibitem{birnbaum2020identifying}
M.~L. Birnbaum, R.~Norel, A.~Van~Meter, A.~F. Ali, E.~Arenare, E.~Eyigoz, C.~Agurto, N.~Germano, J.~M. Kane, and G.~A. Cecchi, ``Identifying signals associated with psychiatric illness utilizing language and images posted to facebook,'' \emph{npj Schizophrenia}, vol.~6, no.~1, pp. 1--10, 2020.

\bibitem{info:doi/10.2196/17650}
G.~Li, B.~Li, L.~Huang, and S.~Hou, ``Automatic construction of a depression-domain lexicon based on microblogs: Text mining study,'' \emph{JMIR Med. Inform.}, vol.~8, no.~6, p. e17650, Jun 2020.

\bibitem{inproceedingsvedula}
N.~Vedula and S.~Parthasarathy, ``Emotional and linguistic cues of depression from social media,'' in \emph{Proc. of Int. Conf. on Digital Health}, 07 2017, pp. 127--136.

\bibitem{rude2004language}
S.~Rude, E.-M. Gortner, and J.~Pennebaker, ``Language use of depressed and depression-vulnerable college students,'' \emph{Cognition \& Emotion}, vol.~18, no.~8, pp. 1121--1133, 2004.

\bibitem{fekete2002internet}
S.~Fekete, ``The internet-a new source of data on suicide, depression and anxiety: a preliminary study,'' \emph{Archives of Suicide Research}, vol.~6, no.~4, pp. 351--361, 2002.

\bibitem{bucur2021exploratory}
A.-M. Bucur, M.~Zampieri, and L.~P. Dinu, ``An exploratory analysis of the relation between offensive language and mental health,'' in \emph{Findings of ACL-IJCNLP 2021}, 2021, pp. 3600--3606.

\bibitem{yates-etal-2017-depression}
A.~Yates, A.~Cohan, and N.~Goharian, ``Depression and self-harm risk assessment in online forums,'' in \emph{Proc. of EMNLP}, Sep. 2017.

\bibitem{hamad2021depressionnet}
Z.~Hamad, R.~Imran, M.~S. Jameel, and X.~Guandong, ``Depressionnet: A novel summarization boosted deep framework for depression detection on social media,'' in \emph{Proc. of SIGIR}.\hskip 1em plus 0.5em minus 0.4em\relax ACM, 2021, pp. 133--142.

\bibitem{10.1007/s10844-020-00599-5}
C.~Y. Chiu, H.~Y. Lane, J.~L. Koh, and A.~L.~P. Chen, ``Multimodal depression detection on instagram considering time interval of posts,'' \emph{J. Intell. Inf. Syst.}, vol.~56, no.~1, p. 25–47, feb 2021.

\bibitem{suhara2017deepmood}
Y.~Suhara, Y.~Xu, and A.~Pentland, ``Deepmood: Forecasting depressed mood based on self-reported histories via recurrent neural networks,'' in \emph{Proc. of ACM Web Conference}, 2017, pp. 715--724.

\bibitem{nanomi-arachchige-etal-2021-dataset-research}
I.~A. Nanomi~Arachchige, V.~H. Jayasuriya, and R.~Weerasinghe, ``A dataset for research on modelling depression severity in online forum data,'' in \emph{Proc. of the SRW, RANLP}.\hskip 1em plus 0.5em minus 0.4em\relax Online: INCOMA Ltd., Sep. 2021, pp. 144--153.

\bibitem{sadeque2018measuring}
F.~Sadeque, D.~Xu, and S.~Bethard, ``Measuring the latency of depression detection in social media,'' in \emph{Proc. of ACM WSDM}, 2018, pp. 495--503.

\bibitem{haque2021deep}
A.~Haque, V.~Reddi, and T.~Giallanza, ``Deep learning for suicide and depression identification with unsupervised label correction,'' in \emph{Artificial Neural Networks and Machine Learning -- ICANN 2021}, I.~Farka{\v{s}}, P.~Masulli, S.~Otte, and S.~Wermter, Eds.\hskip 1em plus 0.5em minus 0.4em\relax Cham: Springer International Publishing, 2021, pp. 436--447.

\bibitem{bucur2021early}
A.-M. Bucur, A.~Cosma, and L.~P. Dinu, ``Early risk detection of pathological gambling, self-harm and depression using bert,'' in \emph{CLEF (Working Notes)}, 2021.

\bibitem{wolk2021hybrid}
A.~Wołk, K.~Chlasta, and P.~Holas, ``Hybrid approach to detecting symptoms of depression in social media entries,'' 2021.

\bibitem{podina14mental}
I.~R. Podina, A.-M. Bucur, D.~Todea, L.~A. Fodor, A.~I. Luca, L.~Dinu, and R.~Boian, ``Mental health at different stages of cancer survival: A natural language processing study of reddit posts,'' \emph{Frontiers in Psychology}, vol.~14, p. 1150227, 2023.

\bibitem{dinu2021automatic}
A.~Dinu and A.-C. Moldovan, ``Automatic detection and classification of mental illnesses from general social media texts,'' in \emph{Proc. of RANLP}, 2021, pp. 358--366.

\bibitem{DBLP:journals/corr/abs-2102-09427}
A.~Haque, V.~Reddi, and T.~Giallanza, ``Deep learning for suicide and depression identification with unsupervised label correction,'' \emph{CoRR}, vol. abs/2102.09427, 2021.

\bibitem{Wu2021ARM}
S.-H. Wu and Z.~Qiu, ``A roberta-based model on measuring the severity of the signs of depression,'' in \emph{Proc. of CLEF}, 2021.

\bibitem{bucur2022end}
A.-M. Bucur, A.~Cosma, L.~P. Dinu, and P.~Rosso, ``An end-to-end set transformer for user-level classification of depression and gambling disorder,'' \emph{CLEF (Working Notes)}, 2022.

\bibitem{ji2022mentalbert}
S.~Ji, T.~Zhang, L.~Ansari, J.~Fu, P.~Tiwari, and E.~Cambria, ``Mentalbert: Publicly available pretrained language models for mental healthcare,'' in \emph{Proc. of LREC}, 2022, pp. 7184--7190.

\bibitem{ji2023domain}
S.~Ji, T.~Zhang, K.~Yang, S.~Ananiadou, E.~Cambria, and J.~Tiedemann, ``Domain-specific continued pretraining of language models for capturing long context in mental health,'' \emph{arXiv preprint arXiv:2304.10447}, 2023.

\bibitem{naseem2022benchmarking}
U.~Naseem, B.~C. Lee, M.~Khushi, J.~Kim, and A.~Dunn, ``Benchmarking for public health surveillance tasks on social media with a domain-specific pretrained language model,'' in \emph{Proc. of NLP Power! Workshop}, 2022, pp. 22--31.

\bibitem{aragon2023disorbert}
M.~Aragon, A.~P.~L. Monroy, L.~Gonzalez, D.~E. Losada, and M.~Montes, ``Disorbert: A double domain adaptation model for detecting signs of mental disorders in social media,'' in \emph{Proc. of ACL}, 2023, pp. 15\,305--15\,318.

\bibitem{xu2023leveraging}
X.~Xu, B.~Yao, Y.~Dong, H.~Yu, J.~Hendler, A.~K. Dey, and D.~Wang, ``Leveraging large language models for mental health prediction via online text data,'' \emph{arXiv preprint arXiv:2307.14385}, 2023.

\bibitem{yang2023evaluations}
K.~Yang, S.~Ji, T.~Zhang, Q.~Xie, and S.~Ananiadou, ``Towards interpretable mental health analysis with chatgpt,'' \emph{arXiv preprint arXiv:2304.03347}, 2023.

\bibitem{qin2023read}
W.~Qin, Z.~Chen, L.~Wang, Y.~Lan, W.~Ren, and R.~Hong, ``Read, diagnose and chat: Towards explainable and interactive llms-augmented depression detection in social media,'' \emph{arXiv preprint arXiv:2305.05138}, 2023.

\bibitem{amin38will}
M.~M. Amin, E.~Cambria, and B.~W. Schuller, ``Will affective computing emerge from foundation models and general ai? a first evaluation on chatgpt,'' \emph{IEEE Intelligent Systems}, vol.~38, p.~2, 2023.

\bibitem{perez2023depresym}
A.~P{\'e}rez, M.~Fern{\'a}ndez-Pichel, J.~Parapar, and D.~E. Losada, ``Depresym: A depression symptom annotated corpus and the role of llms as assessors of psychological markers,'' \emph{arXiv preprint arXiv:2308.10758}, 2023.

\bibitem{bucur2023utilizing}
A.-M. Bucur, ``Utilizing chatgpt generated data to retrieve depression symptoms from social media,'' \emph{CLEF (Working Notes)}, 2023.

\bibitem{souto2023explainability}
E.~B. Souto, A.~P{\'e}rez, and J.~Parapar, ``Explainability, interpretability, depression detection, social media,'' \emph{arXiv preprint arXiv:2310.13664}, 2023.

\bibitem{yang2023mentalllama}
K.~Yang, T.~Zhang, Z.~Kuang, Q.~Xie, and S.~Ananiadou, ``Mentalllama: Interpretable mental health analysis on social media with large language models,'' \emph{arXiv preprint arXiv:2309.13567}, 2023.

\bibitem{taori2023stanford}
R.~Taori, I.~Gulrajani, T.~Zhang, Y.~Dubois, X.~Li, C.~Guestrin, P.~Liang, and T.~B. Hashimoto, ``Stanford alpaca: An instruction-following llama model,'' 2023.

\bibitem{chung2022scaling}
H.~W. Chung, L.~Hou, S.~Longpre, B.~Zoph, Y.~Tay, W.~Fedus, E.~Li, X.~Wang, M.~Dehghani, S.~Brahma \emph{et~al.}, ``Scaling instruction-finetuned language models,'' \emph{arXiv preprint arXiv:2210.11416}, 2022.

\bibitem{openai2022chatgpt}
OpenAI, ``Introducing chatgpt,'' \emph{arXiv}, 2022.

\bibitem{bubeck2023sparks}
S.~Bubeck, V.~Chandrasekaran, R.~Eldan, J.~Gehrke, E.~Horvitz, E.~Kamar, P.~Lee, Y.~T. Lee, Y.~Li, S.~Lundberg \emph{et~al.}, ``Sparks of artificial general intelligence: Early experiments with gpt-4,'' \emph{arXiv preprint arXiv:2303.12712}, 2023.

\bibitem{chiang2023vicuna}
W.-L. Chiang, Z.~Li, Z.~Lin, Y.~Sheng, Z.~Wu, H.~Zhang, L.~Zheng, S.~Zhuang, Y.~Zhuang, J.~E. Gonzalez \emph{et~al.}, ``Vicuna: An open-source chatbot impressing gpt-4 with 90\%* chatgpt quality,'' \emph{See https://vicuna. lmsys. org (accessed 14 April 2023)}, 2023.

\bibitem{bankston2022study}
J.~F. Bankston and L.~Ma, ``A study on people's mental health on twitter during the covid-19 pandemic,'' in \emph{Proc. of ICCDA}, 2022, pp. 111--115.

\bibitem{tommasel2022tracking}
A.~Tommasel, A.~Diaz-Pace, D.~Godoy, and J.~M. Rodriguez, ``Tracking the evolution of crisis processes and mental health on social media during the covid-19 pandemic,'' \emph{Behaviour \& Information Technology}, vol.~41, no.~16, pp. 3450--3469, 2022.

\bibitem{cohrdes2021indications}
C.~Cohrdes, S.~Yenikent, J.~Wu, B.~Ghanem, M.~Franco-Salvador, F.~Vogelgesang \emph{et~al.}, ``Indications of depressive symptoms during the covid-19 pandemic in germany: comparison of national survey and twitter data,'' \emph{JMIR mental health}, vol.~8, no.~6, p. e27140, 2021.

\bibitem{biester2020quantifying}
L.~Biester, K.~Matton, J.~Rajendran, E.~M. Provost, and R.~Mihalcea, ``Quantifying the effects of covid-19 on mental health support forums,'' in \emph{Proc. of NLP-COVID19 Workshop, EMNLP}, 2020.

\bibitem{fine2020assessing}
A.~Fine, P.~Crutchley, J.~Blase, J.~Carroll, and G.~Coppersmith, ``Assessing population-level symptoms of anxiety, depression, and suicide risk in real time using nlp applied to social media data,'' in \emph{Proc. of NLP+CSS Workshop}, 2020, pp. 50--54.

\bibitem{tabak2020temporal}
T.~Tabak and M.~Purver, ``Temporal mental health dynamics on social media,'' in \emph{Proc. of NLP-COVID19 Workshop, EMNLP}, 2020.

\bibitem{zhou2021detecting}
J.~Zhou, H.~Zogan, S.~Yang, S.~Jameel, G.~Xu, and F.~Chen, ``Detecting community depression dynamics due to covid-19 pandemic in australia,'' \emph{IEEE TCSS}, vol.~8, no.~4, pp. 982--991, 2021.

\bibitem{li2020we}
I.~Li, Y.~Li, T.~Li, S.~Alvarez-Napagao, D.~Garcia-Gasulla, and T.~Suzumura, ``What are we depressed about when we talk about covid-19: Mental health analysis on tweets using natural language processing,'' in \emph{Int. Conf. on Innov. Tech. and App. of AI}, 2020, pp. 358--370.

\bibitem{zhang2021monitoring}
Y.~Zhang, H.~Lyu, Y.~Liu, X.~Zhang, Y.~Wang, and J.~Luo, ``Monitoring depression trends on twitter during the covid-19 pandemic: Observational study,'' \emph{JMIR infodemiology}, vol.~1, no.~1, p. e26769, 2021.

\bibitem{biester2021understanding}
L.~Biester, K.~Matton, J.~Rajendran, E.~M. Provost, and R.~Mihalcea, ``Understanding the impact of covid-19 on online mental health forums,'' \emph{ACM Transactions on Management Information Systems}, vol.~12, no.~4, pp. 1--28, 2021.

\bibitem{chen2022covid}
J.~Q. Chen, K.~Qi, A.~Zhang, M.~Shalaginov, and T.~H. Zeng, ``Covid-19 impact on mental health analysis based on reddit comments,'' in \emph{Proc. of BIBM}.\hskip 1em plus 0.5em minus 0.4em\relax IEEE, 2022, pp. 2253--2258.

\bibitem{wolohan-etal-2018-detecting}
J.~Wolohan, M.~Hiraga, A.~Mukherjee, Z.~A. Sayyed, and M.~Millard, ``Detecting linguistic traces of depression in topic-restricted text: Attending to self-stigmatized depression with {NLP},'' in \emph{Proc. of LCCM Workshop}.\hskip 1em plus 0.5em minus 0.4em\relax Santa Fe, New Mexico, USA: ACL, Aug. 2018, pp. 11--21.

\bibitem{bajaj2021mental}
G.~S. Bajaj, H.~Yadav, H.~S. Sahdev, S.~Sah, and P.~Kaur, ``Mental health analysis during covid-19: A comparison before and during the pandemic,'' in \emph{Proc. of IEEE GUCON}.\hskip 1em plus 0.5em minus 0.4em\relax IEEE, 2021, pp. 1--7.

\bibitem{sukhwal2022determining}
P.~C. Sukhwal and A.~Kankanhalli, ``Determining containment policy impacts on public sentiment during the pandemic using social media data,'' \emph{PNAS}, vol. 119, no.~19, p. e2117292119, 2022.

\bibitem{FernandezBarrera2022evaluating}
I.~Fernández-Barrera, S.~Bravo-Bustos, and M.~Vidal, ``Evaluating the social media users' mental health status during covid-19 pandemic using deep learning,'' in \emph{International Conference on Biomedical and Health Informatics}, 2022.

\bibitem{billings2021experiences}
J.~Billings, B.~C.~F. Ching, V.~Gkofa, T.~Greene, and M.~Bloomfield, ``Experiences of frontline healthcare workers and their views about support during covid-19 and previous pandemics: a systematic review and qualitative meta-synthesis,'' \emph{BMC health services research}, vol.~21, pp. 1--17, 2021.

\bibitem{li2020modeling}
D.~Li, H.~Chaudhary, and Z.~Zhang, ``Modeling spatiotemporal pattern of depressive symptoms caused by covid-19 using social media data mining,'' \emph{Int. J. Environ. Res. Public Health}, vol.~17, no.~14, p. 4988, 2020.

\bibitem{saha2020psychosocial}
K.~Saha, J.~Torous, E.~D. Caine, and M.~De~Choudhury, ``Psychosocial effects of the covid-19 pandemic: large-scale quasi-experimental study on social media,'' \emph{JMIR}, vol.~22, no.~11, p. e22600, 2020.

\bibitem{zhang2022covid}
S.~Zhang, L.~Sun, D.~Zhang, P.~Li, Y.~Liu, A.~Anand, Z.~Xie, and D.~Li, ``The covid-19 pandemic and mental health concerns on twitter in the united states,'' \emph{Health data science}, vol. 2022, 2022.

\bibitem{li2022tracking}
M.~Li, Y.~Hua, Y.~Liao, L.~Zhou, X.~Li, L.~Wang, and J.~Yang, ``Tracking the impact of covid-19 and lockdown policies on public mental health using social media: Infoveillance study,'' \emph{JMIR}, vol.~24, no.~10, p. e39676, 2022.

\bibitem{levanti2023depression}
D.~Levanti, R.~N. Monastero, M.~Zamani, J.~C. Eichstaedt, S.~Giorgi, H.~A. Schwartz, and J.~R. Meliker, ``Depression and anxiety on twitter during the covid-19 stay-at-home period in 7 major us cities,'' \emph{AJPM focus}, vol.~2, no.~1, p. 100062, 2023.

\bibitem{tshimula2022covid}
J.~M. Tshimula, B.~Chikhaoui, and S.~Wang, ``Covid-19: Detecting depression signals during stay-at-home period,'' \emph{Health Informatics Journal}, vol.~28, no.~2, p. 14604582221094931, 2022.

\bibitem{marshall2022using}
C.~Marshall, K.~Lanyi, R.~Green, G.~C. Wilkins, F.~Pearson, D.~Craig \emph{et~al.}, ``Using natural language processing to explore mental health insights from uk tweets during the covid-19 pandemic: infodemiology study,'' \emph{Jmir Infodemiology}, vol.~2, no.~1, p. e32449, 2022.

\bibitem{bashar2022tracking}
M.~K. Bashar, ``Tracking public depression from tweets on covid-19 and its comparison with pre-pandemic time,'' in \emph{Proc. of ICIIBMS}, vol.~7.\hskip 1em plus 0.5em minus 0.4em\relax IEEE, 2022, pp. 389--395.

\bibitem{gour2021depression}
G.~Gour, V.~S. Savantanavar, Yashoda, V.~Gadyal, and S.~Basavaraddi, ``Depression analysis of real time tweets during covid pandemic,'' in \emph{Proc. of ICUIS}.\hskip 1em plus 0.5em minus 0.4em\relax Springer, 2021, pp. 55--73.

\bibitem{davis2020quantifying}
B.~D. Davis, D.~E. McKnight, D.~Teodorescu, A.~Quan-Haase, R.~Chunara, A.~Fyshe, and D.~J. Lizotte, ``Quantifying depression-related language on social media during the covid-19 pandemic,'' \emph{International journal of population data science}, vol.~5, no.~4, 2020.

\bibitem{el2021mental}
O.~El-Gayar, A.~Wahbeh, T.~Nasralah, A.~Elnoshokaty, and A.-R. Mohammad, ``Mental health and the covid-19 pandemic: Analysis of twitter discourse,'' \emph{Proc. of AMCIS}, 2021.

\bibitem{nandy2021my}
S.~Nandy and V.~Kumar, ``My mind is a prison: A boosted deep learning approach to detect the rise in depression since covid-19 using a stacked bi-lstm catboost model,'' in \emph{Proc. of Big Data}.\hskip 1em plus 0.5em minus 0.4em\relax IEEE, 2021, pp. 4396--4400.

\bibitem{wu2023exploring}
J.~Wu, X.~Wu, Y.~Hua, S.~Lin, Y.~Zheng, and J.~Yang, ``Exploring social media for early detection of depression in covid-19 patients,'' in \emph{Proc. of ACM Web Conference}, 2023, pp. 3968--3977.

\bibitem{bello2022prevalence}
U.~M. Bello, P.~Kannan, M.~Chutiyami, D.~Salihu, A.~M. Cheong, T.~Miller, J.~W. Pun, A.~S. Muhammad, F.~A. Mahmud, H.~A. Jalo \emph{et~al.}, ``Prevalence of anxiety and depression among the general population in africa during the covid-19 pandemic: a systematic review and meta-analysis,'' \emph{Frontiers in public health}, vol.~10, p. 814981, 2022.

\bibitem{wang2022monitoring}
H.~Wang, L.~Ma, and Y.~Yu, ``Monitoring worldwide trends of expressed depression on twitter before and after covid-19 vaccine releases,'' in \emph{Proc. of ICBCB}.\hskip 1em plus 0.5em minus 0.4em\relax IEEE, 2022, pp. 148--153.

\bibitem{eisenstein2011sparse}
J.~Eisenstein, A.~Ahmed, and E.~P. Xing, ``Sparse additive generative models of text,'' in \emph{Proc. of ICML}, 2011, pp. 1041--1048.

\bibitem{spoorthy2020mental}
M.~S. Spoorthy, S.~K. Pratapa, and S.~Mahant, ``Mental health problems faced by healthcare workers due to the covid-19 pandemic--a review,'' \emph{Asian journal of psychiatry}, vol.~51, p. 102119, 2020.

\bibitem{agarwal2023investigating}
A.~K. Agarwal, J.~Mittal, A.~Tran, R.~Merchant, and S.~C. Guntuku, ``Investigating social media to evaluate emergency medicine physicians’ emotional well-being during covid-19,'' \emph{JAMA Network Open}, vol.~6, no.~5, pp. e2\,312\,708--e2\,312\,708, 2023.

\bibitem{cha2022lexicon}
J.~Cha, S.~Kim, and E.~Park, ``A lexicon-based approach to examine depression detection in social media: the case of twitter and university community,'' \emph{Humanities and Social Sciences Communications}, vol.~9, no.~1, pp. 1--10, 2022.

\bibitem{zogan2023hierarchical}
H.~Zogan, I.~Razzak, S.~Jameel, and G.~Xu, ``Hierarchical convolutional attention network for depression detection on social media and its impact during pandemic,'' \emph{IEEE Journal of Biomedical and Health Informatics}, 2023.

\bibitem{anbalagan2022detecting}
B.~Anbalagan, ``Detecting depressive online user behavior during global pandemic by fusing lstm and cnn models,'' in \emph{Proc. of ICAIAA}.\hskip 1em plus 0.5em minus 0.4em\relax Springer, 2022, pp. 1--10.

\bibitem{meena2022depression}
R.~Meena and V.~T. Bai, ``Depression detection on covid 19 tweets using chimp optimization algorithm.'' \emph{Intelligent Automation \& Soft Computing}, vol.~34, no.~3, 2022.

\bibitem{al2023hybrid}
M.~H. Al~Banna, T.~Ghosh, M.~J.~A. Nahian, M.~S. Kaiser, M.~Mahmud, K.~A. Taher, M.~S. Hossain, and K.~Andersson, ``A hybrid deep learning model to predict the impact of covid-19 on mental health from social media big data,'' \emph{IEEE Access}, 2023.

\bibitem{harrigian2022problem}
K.~Harrigian and M.~Dredze, ``The problem of semantic shift in longitudinal monitoring of social media: A case study on mental health during the covid-19 pandemic,'' in \emph{Proc. of ACM Web Science Conference}, 2022, pp. 208--218.

\bibitem{articlesdht}
B.~Davis, D.~McKnight, D.~Teodorescu, A.~Quan-Haase, R.~Chunara, A.~Fyshe, and D.~Lizotte, ``Quantifying depression-related language on social media during the covid-19 pandemic,'' \emph{International Journal of Population Data Science}, vol.~5, 03 2022.

\bibitem{inproceedingsds}
I.~Fernández-Barrera, S.~Bravo-Bustos, and M.~Vidal, ``Evaluating the social media users' mental health status during covid-19 pandemic using deep learning,'' in \emph{Proc. of ICBHI}, 11 2022.

\bibitem{amir2019mental}
S.~Amir, M.~Dredze, and J.~W. Ayers, ``Mental health surveillance over social media with digital cohorts,'' in \emph{Proc. of CLPsych Workshop, NAACL}, 2019, pp. 114--120.

\bibitem{ishihara2022semantic}
S.~Ishihara, H.~Takahashi, and H.~Shirai, ``Semantic shift stability: Efficient way to detect performance degradation of word embeddings and pre-trained language models,'' in \emph{Proc. of AACL and IJCNLP}, 2022, pp. 205--216.

\bibitem{zhang-etal-2022-symptom}
Z.~Zhang, S.~Chen, M.~Wu, and K.~Zhu, ``Symptom identification for interpretable detection of multiple mental disorders on social media,'' in \emph{Proc. of EMNLP}.\hskip 1em plus 0.5em minus 0.4em\relax Abu Dhabi, United Arab Emirates: ACL, Dec. 2022, pp. 9970--9985.

\bibitem{nguyen2022improving}
T.~Nguyen, A.~Yates, A.~Zirikly, B.~Desmet, and A.~Cohan, ``Improving the generalizability of depression detection by leveraging clinical questionnaires,'' in \emph{Proc. of ACL}, 2022, pp. 8446--8459.

\bibitem{perez2023bdi}
A.~P{\'e}rez, J.~Parapar, {\'A}.~Barreiro, and S.~Lopez-Larrosa, ``Bdi-sen: A sentence dataset for clinical symptoms of depression,'' in \emph{Proc. of ACM SIGIR}, 2023, pp. 2996--3006.

\bibitem{uban2021emotion}
A.-S. Uban, B.~Chulvi, and P.~Rosso, ``An emotion and cognitive based analysis of mental health disorders from social media data,'' \emph{Future Generation Computer Systems}, vol. 124, pp. 480--494, 2021.

\bibitem{ghosh2023attention}
T.~Ghosh, M.~H. Al~Banna, M.~J. Al~Nahian, M.~N. Uddin, M.~S. Kaiser, and M.~Mahmud, ``An attention-based hybrid architecture with explainability for depressive social media text detection in bangla,'' \emph{Expert Systems with Applications}, vol. 213, p. 119007, 2023.

\bibitem{lundberg2017unified}
S.~M. Lundberg and S.-I. Lee, ``A unified approach to interpreting model predictions,'' \emph{Advances in neural information processing systems}, vol.~30, 2017.

\bibitem{ribeiro2016should}
M.~T. Ribeiro, S.~Singh, and C.~Guestrin, ``" why should i trust you?" explaining the predictions of any classifier,'' in \emph{Proc. of ACM SIGKDDg}, 2016, pp. 1135--1144.

\bibitem{ADARSH2023103168}
V.~Adarsh, P.~{Arun Kumar}, V.~Lavanya, and G.~Gangadharan, ``Fair and explainable depression detection in social media,'' \emph{IPM}, vol.~60, no.~1, p. 103168, 2023.

\bibitem{desai2020calibration}
S.~Desai and G.~Durrett, ``Calibration of pre-trained transformers,'' in \emph{Proc. of EMNLP}, 2020, pp. 295--302.

\bibitem{ulmer2022exploring}
D.~Ulmer, J.~Frellsen, and C.~Hardmeier, ``Exploring predictive uncertainty and calibration in nlp: A study on the impact of method \& data scarcity,'' in \emph{Findings of EMNLP 2022}, 2022, pp. 2707--2735.

\bibitem{kenton2019bert}
J.~D. M.-W.~C. Kenton and L.~K. Toutanova, ``Bert: Pre-training of deep bidirectional transformers for language understanding,'' in \emph{Proc. of NAACL-HLT}, 2019, pp. 4171--4186.

\bibitem{liu2019roberta}
Y.~Liu, M.~Ott, N.~Goyal, J.~Du, M.~Joshi, D.~Chen, O.~Levy, M.~Lewis, L.~Zettlemoyer, and V.~Stoyanov, ``Roberta: A robustly optimized bert pretraining approach,'' \emph{arXiv preprint arXiv:1907.11692}, 2019.

\bibitem{ilias2023calibration}
L.~Ilias, S.~Mouzakitis, and D.~Askounis, ``Calibration of transformer-based models for identifying stress and depression in social media,'' \emph{IEEE TCSS}, 2023.

\bibitem{ajmani2023systematic}
L.~H. Ajmani, S.~Chancellor, B.~Mehta, C.~Fiesler, M.~Zimmer, and M.~De~Choudhury, ``A systematic review of ethics disclosures in predictive mental health research,'' in \emph{Proc. of ACM FAccT}, 2023, pp. 1311--1323.

\bibitem{birnbaum2017collaborative}
M.~L. Birnbaum, S.~K. Ernala, A.~F. Rizvi, M.~De~Choudhury, and J.~M. Kane, ``A collaborative approach to identifying social media markers of schizophrenia by employing machine learning and clinical appraisals,'' \emph{JMIR}, vol.~19, no.~8, p. e7956, 2017.

\bibitem{ayers2018don}
J.~W. Ayers, T.~L. Caputi, C.~Nebeker, and M.~Dredze, ``Don’t quote me: reverse identification of research participants in social media studies,'' \emph{NPJ digital medicine}, vol.~1, no.~1, p.~30, 2018.

\bibitem{benton2017ethical}
A.~Benton, G.~Coppersmith, and M.~Dredze, ``Ethical research protocols for social media health research,'' in \emph{Proc. of EthNLP Workshop, ACL}, 2017, pp. 94--102.

\bibitem{mitchell2019model}
M.~Mitchell, S.~Wu, A.~Zaldivar, P.~Barnes, L.~Vasserman, B.~Hutchinson, E.~Spitzer, I.~D. Raji, and T.~Gebru, ``Model cards for model reporting,'' in \emph{Proc. of ACM FAccT}, 2019, pp. 220--229.

\bibitem{boyd2021datasheets}
K.~L. Boyd, ``Datasheets for datasets help ml engineers notice and understand ethical issues in training data,'' \emph{Proc. of the ACM on HCI}, vol.~5, no. CSCW2, pp. 1--27, 2021.

\bibitem{benton2017multitask}
A.~Benton, M.~Mitchell, D.~Hovy \emph{et~al.}, ``Multitask learning for mental health conditions with limited social media data,'' in \emph{Proc. of EACL}.\hskip 1em plus 0.5em minus 0.4em\relax ACL, 2017.

\bibitem{aguirre2021gender}
C.~Aguirre, K.~Harrigian, and M.~Dredze, ``Gender and racial fairness in depression research using social media,'' in \emph{Proc. of EACL}, 2021, pp. 2932--2949.

\bibitem{de2017gender}
M.~De~Choudhury, S.~S. Sharma, T.~Logar, W.~Eekhout, and R.~C. Nielsen, ``Gender and cross-cultural differences in social media disclosures of mental illness,'' in \emph{Proc. of ACM CSCW}, 2017, pp. 353--369.

\bibitem{loveys-etal-2018-cross}
K.~Loveys, J.~Torrez, A.~Fine, G.~Moriarty, and G.~Coppersmith, ``Cross-cultural differences in language markers of depression online,'' in \emph{Proc. of CLPsych Workshop, NAACL}.\hskip 1em plus 0.5em minus 0.4em\relax New Orleans, LA: ACL, Jun. 2018, pp. 78--87.

\bibitem{aguirre2021qualitative}
C.~Aguirre and M.~Dredze, ``Qualitative analysis of depression models by demographics,'' in \emph{Proc. of CLPsych Workshop, NAACL}, 2021, pp. 169--180.

\bibitem{lin2022gendered}
I.~Lin, L.~Njoo, A.~Field, A.~Sharma, K.~Reinecke, T.~Althoff, and Y.~Tsvetkov, ``Gendered mental health stigma in masked language models,'' in \emph{Proc. of EMNLP}, 2022, pp. 2152--2170.

\bibitem{articleeich}
J.~Eichstaedt, R.~Smith, R.~Merchant, L.~Ungar, P.~Crutchley, D.~Preotiuc-Pietro, D.~Asch, and H.~Schwartz, ``Facebook language predicts depression in medical records,'' \emph{PNAS}, vol. 115, p. 201802331, 10 2018.

\bibitem{kelly2020blinded}
D.~L. Kelly, M.~Spaderna, V.~Hodzic, S.~Nair, C.~Kitchen, A.~E. Werkheiser, M.~M. Powell, F.~Liu, G.~Coppersmith, S.~Chen \emph{et~al.}, ``Blinded clinical ratings of social media data are correlated with in-person clinical ratings in participants diagnosed with either depression, schizophrenia, or healthy controls,'' \emph{Psychiatry Research}, vol. 294, p. 113496, 2020.

\bibitem{workshop2022bloom}
B.~Workshop, T.~L. Scao, A.~Fan, C.~Akiki, E.~Pavlick, S.~Ili{\'c}, D.~Hesslow, R.~Castagn{\'e}, A.~S. Luccioni, F.~Yvon \emph{et~al.}, ``Bloom: A 176b-parameter open-access multilingual language model,'' \emph{arXiv preprint arXiv:2211.05100}, 2022.

\bibitem{touvron2023llama}
H.~Touvron, T.~Lavril, G.~Izacard, X.~Martinet, M.-A. Lachaux, T.~Lacroix, B.~Rozi{\`e}re, N.~Goyal, E.~Hambro, F.~Azhar \emph{et~al.}, ``Llama: Open and efficient foundation language models,'' \emph{arXiv preprint arXiv:2302.13971}, 2023.

\bibitem{bucur2023matter}
A.-M. Bucur, A.~Cosma, P.~Rosso, and L.~P. Dinu, ``It's just a matter of time: Detecting depression with time-enriched multimodal transformers,'' in \emph{Advances in Information Retrieval}, J.~Kamps, L.~Goeuriot, F.~Crestani, M.~Maistro, H.~Joho, B.~Davis, C.~Gurrin, U.~Kruschwitz, and A.~Caputo, Eds.\hskip 1em plus 0.5em minus 0.4em\relax Cham: Springer Nature Switzerland, 2023, pp. 200--215.

\bibitem{yadav2023towards}
S.~Yadav, C.~Caragea, C.~Zhao, N.~Kumari, M.~Solberg, and T.~Sharma, ``Towards identifying fine-grained depression symptoms from memes,'' in \emph{Proc. of ACL}, 2023, pp. 8890--8905.

\bibitem{gimeno2024reading}
D.~Gimeno-G{\'o}mez, A.-M. Bucur, A.~Cosma, C.-D. Mart{\'\i}nez-Hinarejos, and P.~Rosso, ``Reading between the frames: Multi-modal depression detection in videos from non-verbal cues,'' \emph{arXiv preprint arXiv:2401.02746}, 2024.

\bibitem{openai2023gpt}
OpenAI, ``Gpt-4 technical report,'' \emph{arXiv}, 2023.

\bibitem{liu2023visual}
H.~Liu, C.~Li, Q.~Wu, and Y.~J. Lee, ``Visual instruction tuning,'' \emph{arXiv preprint arXiv:2304.08485}, 2023.

\bibitem{pires2019multilingual}
T.~Pires, E.~Schlinger, and D.~Garrette, ``How multilingual is multilingual bert?'' in \emph{Proc. of ACL}, 2019, pp. 4996--5001.

\bibitem{conneau2020unsupervised}
A.~Conneau, K.~Khandelwal, N.~Goyal, V.~Chaudhary, G.~Wenzek, F.~Guzm{\'a}n, {\'E}.~Grave, M.~Ott, L.~Zettlemoyer, and V.~Stoyanov, ``Unsupervised cross-lingual representation learning at scale,'' in \emph{Proc. of ACL}, 2020, pp. 8440--8451.

\end{thebibliography}

\end{document}